\documentclass[lettersize,journal]{IEEEtran}

\usepackage[utf8]{inputenc} 
\usepackage[T1]{fontenc}    
\usepackage{hyperref}       
\usepackage{url}            
\usepackage{booktabs}       
\usepackage{amsfonts}       
\usepackage{nicefrac}       
\usepackage{microtype}      
\usepackage{tabularx}

\usepackage{amsmath}       
\usepackage{amssymb}       
\usepackage{amsfonts}      
\usepackage{mathtools}     
\usepackage{bm}            
\usepackage{bbm}           
\usepackage{xfrac}

\usepackage{times}         
\usepackage{dsfont}        
\usepackage{xspace}        
\usepackage{soul}
\setul{1.5pt}{1pt}

\usepackage{graphicx}      
\usepackage{wrapfig}       
\usepackage{float}         
\usepackage{stfloats}      
\usepackage{capt-of}       
\usepackage{caption}      
\newcommand{\boldparagraph}[1]{\vspace{0.1cm}\noindent{\bf #1.}}
\usepackage{subfigure}

\usepackage{booktabs}      
\usepackage{multirow}      
\usepackage[flushleft]{threeparttable} 
\usepackage{arydshln}      
\usepackage{adjustbox}
\usepackage[dvipsnames,table,xcdraw]{xcolor} 
\newcolumntype{L}[1]{>{\raggedright\arraybackslash}p{#1}}
\definecolor{mydarkblue}{rgb}{0,0.08,0.45}
\definecolor{mydarkgreen}{RGB}{0, 139, 69}
\definecolor{mygreen2}{RGB}{0, 205, 0}
\definecolor{mybrown}{RGB}{139, 69, 19}
\definecolor{boxblue}{RGB}{79,173,234}
\definecolor{tablepeach}{RGB}{255, 240, 235}
\definecolor{tablepurple}{RGB}{248,235,252}
\definecolor{tableblue}{RGB}{235,241,255}

\usepackage{hyperref}                   
\usepackage[capitalise]{cleveref} 
\usepackage{xurl}                       
\definecolor{citecolor}{HTML}{c03d3e}

\hypersetup{
    colorlinks=true,
    linkcolor=red,
    urlcolor=mydarkgreen,
    citecolor=mydarkgreen,
}

\usepackage[linesnumbered,ruled,vlined]{algorithm2e} 
\crefname{algocf}{alg.}{algs.}
\Crefname{algocf}{Algorithm}{Algorithms}

\usepackage{pifont}  
\usepackage{dsfont}  
\usepackage{lipsum}  
\usepackage{siunitx} 
\usepackage{multicol} 

\usepackage{enumitem}

\newcommand{\methodname}{SMASH\xspace}

\begin{document}

\title{
\methodname: Mastering Scalable Whole-Body Skills for Humanoid
Ping-Pong with Egocentric Vision
}

\author{
Junli Ren$^{\dagger,*}$,
Yinghui Li$^{\dagger,*}$,
Kai Zhang$^{*}$,
Penglin Fu$^{*}$,
Haoran Jiang,
Yixuan Pan,
Guangjun Zeng,
Tao Huang,\\[-1pt]
Weizhong Guo,
Peng Lu,
Tianyu Li,
Jingbo Wang,
Li Chen,
Hongyang Li,
Ping Luo$^{\ddagger}$\\[3pt]
{\normalsize The University of Hong Kong \quad Kinetix AI}\\[2pt]
{\color{gray!65}\normalsize
$^{*}$Co-first author \quad
$^{\dagger}$Project lead \quad
$^{\ddagger}$Corresponding author
}\\[-1pt]
{\texttt{\href{https://mmlab.hk/Smash/}}{\textcolor[rgb]{0.86,0.10,0.45}{https://mmlab.hk/Smash/}}
}
}

\noindent
\twocolumn[{%
\renewcommand\twocolumn[1][]{#1}
\maketitle
\vspace{-25pt}
\begin{center}
    \includegraphics[width=\textwidth]{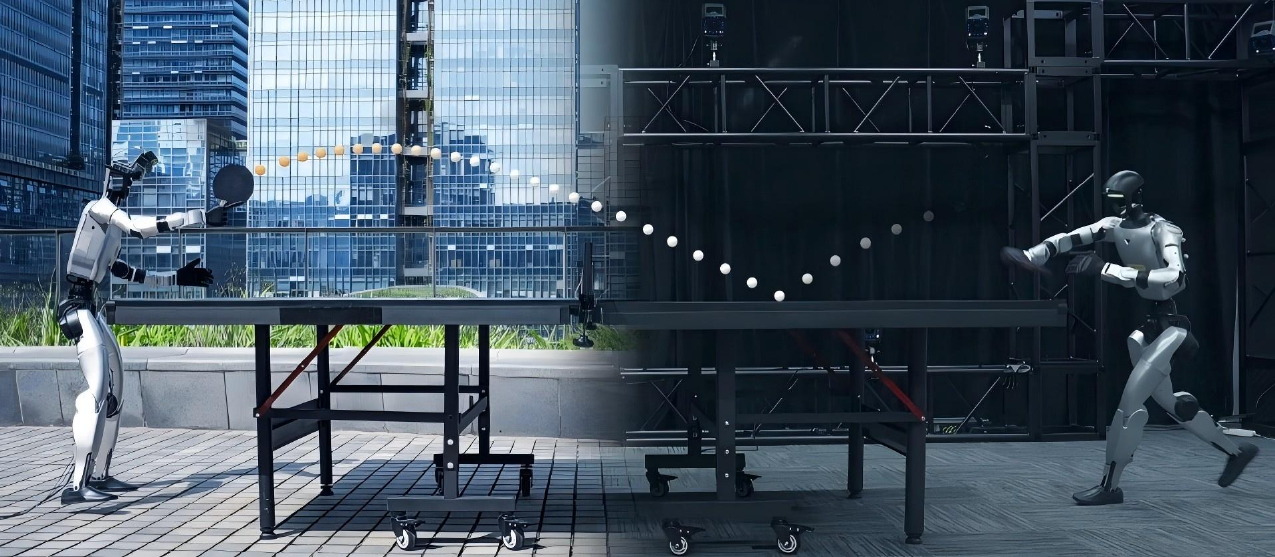}
    \vspace{-10pt}
    \captionof{figure}{\textbf{SMASH}: Our system enables the first outdoor humanoid ping-pong player and the first whole-body smash on a humanoid robot. Through scalable motion generation and whole-body motion matching, the robot achieves expressive and agile ball interaction across a wide hitting workspace.}
    \label{fig:teaser}
    \vspace{-6pt}
\end{center}
}]

\begin{abstract}
Existing humanoid table tennis systems remain limited by their reliance on external sensing and their inability to achieve agile whole-body coordination for precise task execution. 
These limitations stem from two core challenges: achieving low-latency and robust onboard egocentric perception under fast robot motion, and obtaining sufficiently diverse task-aligned strike motions for learning precise yet natural whole-body behaviors. 
In this work, we present \methodname, a modular system for agile humanoid table tennis that unifies scalable whole-body skill learning with onboard egocentric perception, eliminating the need for external cameras during deployment.
Our work advances prior humanoid table-tennis systems in three key aspects.
First, we achieve agile and precise ball interaction with tightly coordinated whole-body control, rather than relying on decoupled upper- and lower-body behaviors. 
This enables the system to exhibit diverse strike motions, including explosive whole-body smashes and low crouching shots.
Second, by augmenting and diversifying strike motions with a generative model, our framework benefits from scalable motion priors and produces natural, robust striking behaviors across a wide workspace. 
Third, to the best of our knowledge, we demonstrate the first humanoid table-tennis system capable of consecutive strikes using onboard sensing alone, despite the challenges of low-latency perception, ego-motion-induced instability, and limited field of view. 
Extensive real-world experiments demonstrate stable and precise ball exchanges under high-speed conditions, validating scalable, perception-driven whole-body skill learning for dynamic humanoid interaction tasks.
\end{abstract}

\section{Introduction}
\label{sec:intro}
How do humans play table tennis? We track the ball with our own eyes, anticipate its future trajectory, and coordinate the whole body to strike at the right position, time, and velocity. Through observation and repeated practice, we also develop a diverse repertoire of strike motions and the prior experience needed for agile response. Endowing humanoid robots with comparable capabilities, however, remains highly challenging, as it requires both agile whole-body striking skills and real-time egocentric onboard perception \cite{Yu2025EgoMILA, ma2025badminton, Kim2025ARMOREP}. Achieving this would bring humanoid robots closer to natural and autonomous interaction in dynamic open-world environments.

We present \methodname, a humanoid table tennis system built around three methodological advances:

First, we develop an egocentric onboard perception pipeline for real-time humanoid table tennis. 
Using only onboard cameras, the system jointly estimates the ball state and the robot pose at over $50\text{Hz}$. 
This setting is particularly challenging because the ping-pong ball is small, fast, and easily blurred \cite{wang2025spikepingpong, DAmbrosio2023RoboticTT}, while the robot must perceive and react within a narrow time window under continuous whole-body motion.

Second, we introduce a task-aligned whole-body control framework based on motion priors. 
Rather than relying on decoupled upper- and lower-body behaviors \cite{su2025hitterhumanoidtabletennis, Ben2025HOMIEHL, Jiang2025WholeBodyVLATU} or a small set of pre-defined motions \cite{wang2026humanx, Han2026HUSKYHS, ren2025goalkeeper}, the framework learns coordinated whole-body striking behaviors that remain both task-effective and physically natural.

Third, we propose a scalable motion generation and matching framework for whole-body striking. 
To address the difficulty of acquiring sufficiently diverse task-aligned strike motions \cite{Wang2024StrategyAS, zhang2026latent, zhang2023broadcasttennis}, we augment motion-capture demonstrations with motions generated by a motion model, yielding a motion library with broad coverage of the reachable hitting workspace. 
We further adopt a simple task-conditioned motion matching scheme to retrieve references aligned with the desired strike target, tightly coupling motion priors with the ball-interaction objective during policy learning.

Taken together, these components establish a new level of capability for humanoid table tennis. 
We demonstrate the first outdoor consecutive humanoid table-tennis striking without external cameras or motion-capture systems \cite{su2025hitterhumanoidtabletennis, wang2026humanx, zhang2026latent}, enabled entirely by egocentric onboard perception. 
Our system further realizes highly dynamic and diverse whole-body behaviors, including smashes, low crouching shots, and large coordinated lateral movements with accurate ball interaction. 
Beyond these system-level results, we show that motion generation provides an effective way to transform sparse and incomplete motion-capture demonstrations into a workspace-covering strike library, substantially improving task--motion matching, training efficiency, and the naturalness of the resulting robot behavior. 
Extensive simulation and real-world experiments validate the proposed approach and demonstrate its robustness, efficiency, and versatility across a wide range of striking scenarios.

\section{Related Work}
\label{sec:relatedwork}

\subsection{Whole-body Control for Robotic Sports}
Recent progress in robotics has moved from basic locomotion \cite{huang2025learning,gu2024advancing} toward agile whole-body behaviors that require rapid contact transitions, dynamic balance recovery, and tight upper- and lower-body coordination \cite{he2025asap,xie2025kungfubot,zhang2024wococo,Liao2025BeyondMimicFM,luo2025sonic,pan2025ams}, with growing attention to interactive ball sports such as basketball~\cite{wang2026humanx}, soccer~\cite{kong2026paid, wang2025reactivesoccer}, tennis~\cite{zhang2026latent}, badminton~\cite{ma2025badminton, Chen2026LearningHB} and table tennis~\cite{su2025hitterhumanoidtabletennis,hu2025pace, Guist2023SafeA, Bchler2020LearningTP, dambrosio2025competitivett}.
Among these interactive ball sports, humanoid table tennis is a representative benchmark that requires coordinated whole-body striking, rapid repositioning, and reliable recovery within extremely limited reaction time.
In humanoid table tennis, HITTER~\cite{su2025hitterhumanoidtabletennis} and PACE~\cite{hu2025pace} exemplify two common design choices: hierarchical planning with upper-body-centered imitation, and end-to-end reinforcement learning with predictive augmentation and physics-guided task rewards, without explicit human motion references or imitation objectives.
\methodname instead combines full-body strike tracking from human motion with task-driven reinforcement learning, enabling more human-like whole-body coordination and diverse striking styles.

\subsection{Onboard Perception for Dynamic Humanoid Interaction}
Perception remains a major bottleneck for deployable humanoid systems in dynamic interaction tasks. 
Recent work has begun to couple onboard perception with humanoid or legged control in scene interaction~\cite{wang2025physhsi,allshire2025videomimic, Ren2025VBComLV, Wang2025BeamDojoLA}, parkour~\cite{wu2026php,zhuang2026deepwholebodyparkour, Long2024LearningHL}, and sports, demonstrating closed-loop visuomotor behavior beyond static settings \cite{wang2025reactivesoccer,ren2025goalkeeper}. 
In ball sports, legged badminton~\cite{ma2025badminton} further shows that autonomous visual perception can be integrated with whole-body control through shuttlecock prediction under real camera noise. 
However, in the more demanding setting of table tennis, prior systems still mainly rely on external sensing, including specialized high-speed vision setups~\cite{wang2025spikepingpong,dambrosio2025competitivett} or motion-capture infrastructure \cite{su2025hitterhumanoidtabletennis,hu2025pace}. 
This difference is significant because table tennis involves a smaller ball, shorter interaction distances, and tighter reaction windows. \methodname fills this gap by enabling continuous humanoid table-tennis interaction using only ego-centric onboard visual sensing and online state estimation.

\subsection{Motion Task Alignment}

Aligning motion priors with task objectives remains a central challenge in imitation- and reinforcement-learning-based whole-body control. 
Existing approaches address this problem from several perspectives, but still exhibit notable limitations.
Adversarial Motion Priors (AMP)~\cite{ren2025goalkeeper, wang2025physhsi} typically rely on a single motion distribution and enforce motion realism through adversarial rewards, often resulting in loose task alignment and limited motion diversity. 
Other works attempt to improve data coverage through motion augmentation or retargeting~\cite{wang2026humanx, weng2025hdmi, Huang2025TowardsAH, yang2025omniretarget, su2025hitterhumanoidtabletennis}, but these approaches are often constrained by the need to preserve motion tracking fidelity, leading to limited task generalization.
Another alternative is two-stage frameworks, where a high-level policy generates task-aligned reference motion and a low-level controller tracks reference motions~\cite{zhang2023broadcasttennis, zhang2026latent}. 
However, such decoupled designs introduce additional training complexity, and instability in high-level outputs can degrade motion quality and increase the sim-to-real gap.
Recent work explores motion matching as a way to leverage large motion datasets by retrieving reference motions based on state or motion similarity~\cite{wu2026php}. 
Different from this formulation, our approach performs motion retrieval conditioned on the desired strike target, allowing the selected references to be directly aligned with the ball-interaction objective during policy learning. 
To address the limited coverage and uneven distribution of motion-capture data, we further augment demonstrations with a generative motion model, yielding a strike-motion library with broad coverage of the reachable hitting workspace. 
This combination enables task-aligned motion priors for precise, natural, and diverse whole-body striking.

\section{\methodname Overview}
\label{sec:method}


\begin{figure*}[t!]
  \centering
  \includegraphics[width= \textwidth]{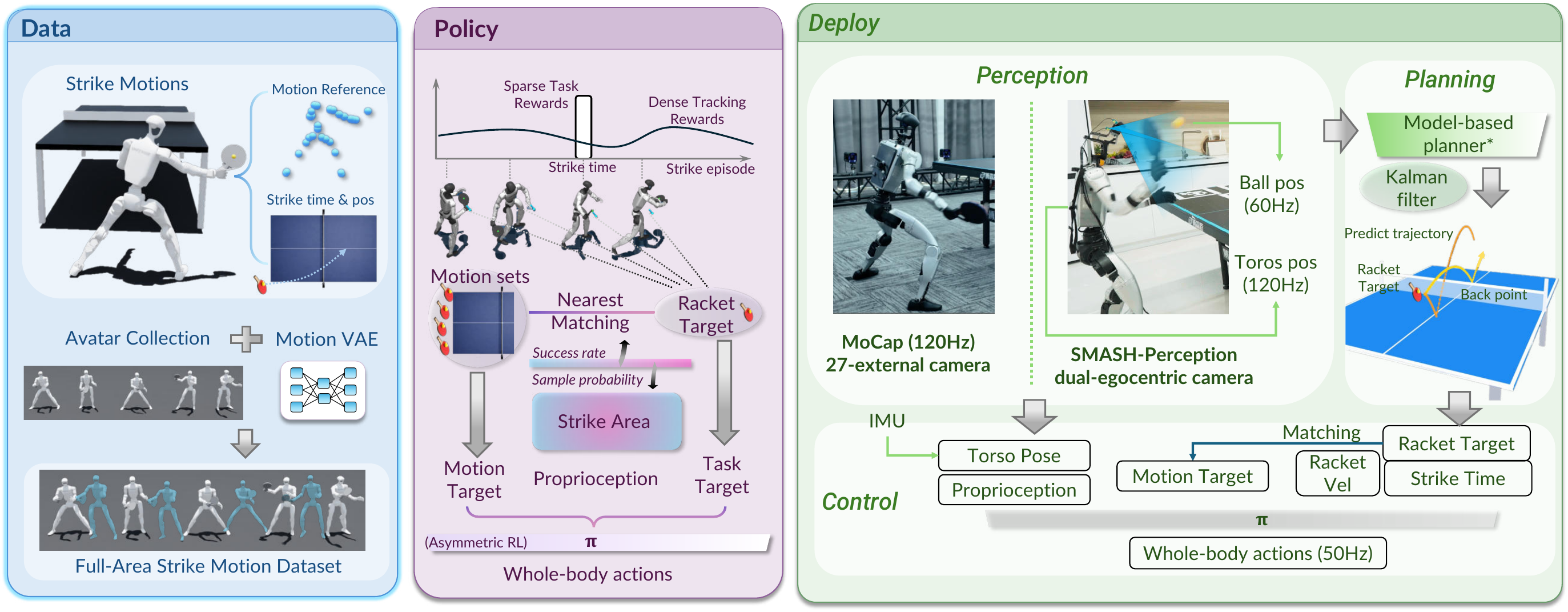} 
\caption{\textbf{Overview of \methodname}. Our system connects scalable motion generation, task-aligned policy learning, and egocentric onboard perception into a unified pipeline for humanoid table tennis. \textbf{Data}: Motion-capture demonstrations are augmented with a motion VAE to build a strike-motion dataset that covers the reachable hitting workspace. \textbf{Policy}: A whole-body policy is trained via reinforcement learning, where task commands are tightly coupled with motion priors through nearest motion matching, enabling the selection and execution of appropriate strike behaviors. \textbf{Deploy}: At test time, egocentric onboard perception provides real-time estimates of ball and robot states, which are used by a planner and matching module to generate closed-loop whole-body actions. This pipeline enables precise and natural ball interaction without relying on external sensing infrastructure.}
  \label{fig:method}
\end{figure*}

\methodname (\cref{fig:method}) is a modular system that achieves agile table tennis behaviour on humanoid robots. 
It distinguishes itself from prior work in two key aspects: task-aligned imitation learning for precise strike control, and egocentric onboard perception for real-time interaction.

First, we construct a strike motion dataset that covers the reachable hitting workspace. 
The dataset combines motion capture demonstrations with motions augmented by a motion variational autoencoder (VAE), expanding strike coverage and enabling scalable motion references. 
Second, we train a whole-body control policy that integrates reinforcement learning with large-scale motion constraints. 
The workspace-covering strike dataset enables motion matching with references aligned with the desired strike target. 
This allows the policy to achieve accurate ball interaction while preserving natural whole-body coordination. 
Third, we build a robust egocentric onboard perception system for real-time ball observation during play. 
The estimated ball state supports downstream planning and control for strike execution.

Together, these components enable agile whole-body table tennis on humanoid robots using onboard sensing. 
The robot is no longer limited to simple forehand and backhand strokes, but can execute diverse whole-body strikes such as smashes and low crouching shots. 
Moreover, the system can operate without external cameras, enabling the first outdoor humanoid table tennis interactions.

\subsection*{\textbf{Preliminaries}}
\paragraph{Motion tracking with reinforcement learning}
Recent humanoid control methods commonly combine reinforcement learning with motion imitation to train policies for robust whole-body motion tracking in simulation and transfer to real hardware \cite{Liao2025BeyondMimicFM, he2025asap, He2024OmniH2OUA}. 
They are typically formulated as a Partially Observable Markov Decision Process (POMDP), where the actor observes deployment-feasible inputs such as motion commands and proprioception, the critic uses additional privileged information during training, and domain randomization is applied to reduce the sim-to-real gap. 
The policy is then optimized with PPO. 
Building on this paradigm, we further introduce task objectives on top of motion-prior tracking. 
The central challenge is therefore to ensure that the selected motion prior matches the current task target.

\paragraph{Model-based strike target planning}
For humanoid ping-pong, a common design is to first estimate the ball state from perception, then predict the future ball trajectory, and finally determine a feasible strike target for the whole-body controller \cite{su2025hitterhumanoidtabletennis, Ko2018OnlineOT, Huang2016JointlyLT}. 
In particular, the observed ball positions are used to estimate the ball velocity, the predicted trajectory determines the desired hitting point and hitting time, and the desired return target is used to plan the racket velocity at contact. 
In our setting, the return objective is chosen as the center of the opponent-side table.

Accordingly, we represent the strike command by
\begin{equation}
\mathbf{y}_{\mathrm{hit}}=\left(\mathbf{p}_{\mathrm{hit}},\mathbf{v}_{\mathrm{hit}},\tau\right),
\end{equation}
where \(\mathbf{p}_{\mathrm{hit}}\) is the desired hitting position, \(\mathbf{v}_{\mathrm{hit}}\) is the desired hitting velocity, and \(\tau\) is the time-to-strike. 
These task-space quantities form the interface between ball-trajectory planning and whole-body policy control in our system.

\section{Learning scalable whole-body skills}
\label{sec:scalable_skill}

A central challenge in humanoid table tennis is that whole-body strike priors must cover a broad hitting workspace, yet motion-capture data are inherently difficult to scale to this requirement.
The limitation is not merely collection cost: dynamic whole-body strikes are hard to reproduce consistently, the feasible hitting space is large, and many target regions are rarely observed even after substantial data collection.
Consequently, the recorded strike motions remain sparse and unevenly distributed across the table. 
When used directly as motion priors, such incomplete coverage leaves many feasible strike targets without suitable references, which degrades motion matching and ultimately limits policy performance.

\begin{figure}[t!]
  \centering
  \includegraphics[width= 0.47\textwidth]{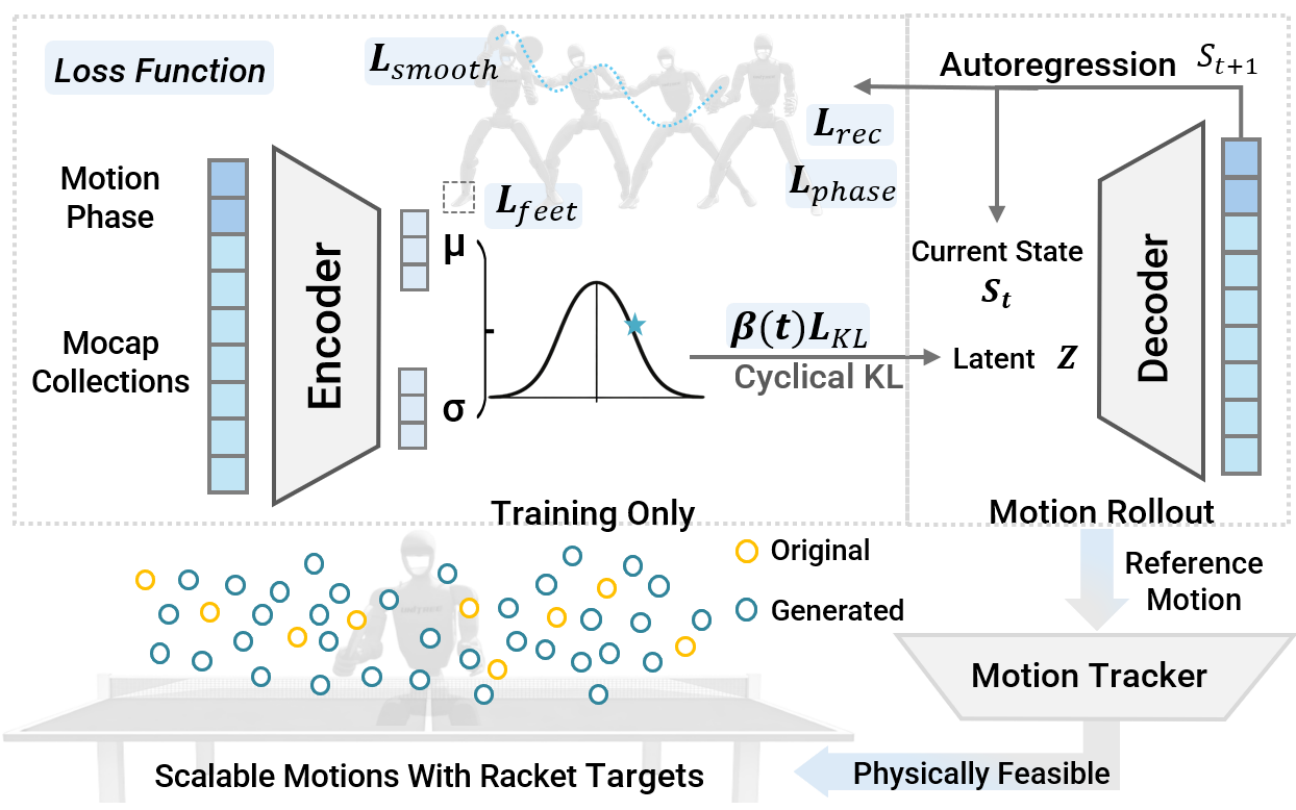} 
   \caption{\textbf{Motion-VAE} for scalable strike motion generation.}
  \label{fig:vae}
\end{figure}

To address this limitation, we construct a scalable strike motion library by augmenting the captured motions with a task-aligned generative model. 
Our goal is not to synthesize arbitrary full-body motions, but to generate strike motions that preserve human-like coordination and timing while expanding coverage of the table-wide hitting workspace. 
To this end, as shown in \cref{fig:vae}, we train a conditional autoregressive Motion-VAE on strike-centered motion clips, and then convert the generated motions into physically executable references through tracker-based rollout and filtering.

\subsection{Task-aligned Strike Motion Generation}
\label{sec:task_aligned_generation}

We begin by collecting 400 motion-capture recordings of whole-body strike demonstrations. During data collection, we annotate only the strike instant for each demonstration. Around this instant, we extract a motion segment of 1.08,s, spanning 0.54,s before and 0.54,s after contact, so that each clip contains a complete strike cycle including preparation, acceleration, contact, and recovery. These human demonstrations are then retargeted to the Unitree G1 kinematics via GMR~\cite{araujo2025gmr}. For each retargeted motion, we further compute a strike target feature defined as the relative position of the racket link at the strike instant $(t=0.54s)$ with respect to the torso link at the initial frame $(t=0s)$. This process yields a dataset of 400 whole-body strike motions, each paired with its corresponding strike target at contact.

Although these demonstrations are natural and high-quality, their associated strike points cover only a limited subset of the reachable hitting workspace. 
We therefore train a task-aligned Motion-VAE to synthesize additional strike motions and improve the coverage of the resulting motion prior.

Let $\mathbf{s}_t \in \mathbb{R}^{d_s}$ denote the motion state at time step $t$, where $\mathbf{s}_t$ includes the kinematic features used to represent the reference motion. 
Given a conditioning context and a latent variable $\mathbf{z}\in\mathbb{R}^{d_z}$, the Motion-VAE predicts future motion states autoregressively. 
Specifically, the encoder defines an approximate posterior
\begin{equation}
q_{\phi}\!\left(\mathbf{z}\mid \mathbf{s}_{t:t+H}\right),
\end{equation}
where the latent variable $\mathbf{z}$ is obtained via the reparameterization trick: $\mathbf{z} = \boldsymbol{\mu}_{\phi} + \boldsymbol{\sigma}_{\phi} \odot \boldsymbol{\epsilon}$, where $\boldsymbol{\epsilon} \sim \mathcal{N}(\mathbf{0}, \mathbf{I})$. And the decoder models future motion generation as
\begin{equation}
p_{\theta}\!\left(\hat{\mathbf{s}}_{t+1:t+H}\mid \mathbf{s}_{t}^{\mathrm{cond}}, \mathbf{z}\right),
\end{equation}
where $\mathbf{s}_{t}^{\mathrm{cond}}$ denotes the conditioning state and $H$ is the rollout horizon.

Following the standard Motion-VAE formulation~\cite{ling2020character, yuan2023learning}, we first optimize a base generative objective that combines motion reconstruction and latent regularization. 
Specifically, the Motion-VAE loss is defined as
\begin{equation}
\mathcal{L}_{\mathrm{VAE}} = \mathcal{L}_{\mathrm{rec}} + \beta(t) \mathcal{L}_{\mathrm{KL}},
\end{equation}
where $\mathcal{L}_{\mathrm{rec}}$ is the multi-step motion reconstruction loss and $\mathcal{L}_{\mathrm{KL}}$ regularizes the approximate posterior toward a Gaussian prior. To prevent posterior collapse throughout VAE training, we adopt a cyclical annealing schedule for the KL weight $\beta(t)$. This allows model to focus on reconstruction in early phase of each cycle while progressively regularizing the latent space as $\beta(t)$ increases.  

While this base objective is sufficient for generic motion modeling, it does not explicitly enforce the temporal structure and physical plausibility required by table-tennis strike motions. 
We therefore augment it with three task-aligned regularization terms.

First, we introduce a phase reconstruction loss to preserve the strike-cycle structure. 
Each frame is annotated with a cyclic phase representation
\begin{equation}
\mathbf{c}_t
=
\begin{bmatrix}
\sin \varphi_t \\[2pt]
\cos \varphi_t
\end{bmatrix},
\end{equation}
where $\varphi_t$ denotes the strike phase. 
The predicted phase is supervised by
\begin{equation}
\mathcal{L}_{\mathrm{phase}}
=
\sum_{h=1}^{H}
\left\|
\hat{\mathbf{c}}_{t+h}
-
\mathbf{c}_{t+h}
\right\|_2^2
+
\sum_{h=1}^{H-1}
\left\|
\Delta\hat{\mathbf{c}}_{t+h}
-
\Delta\mathbf{c}_{t+h}
\right\|_2^2,
\end{equation}
where $\Delta\mathbf{c}_{t+h} = \mathbf{c}_{t+h+1} - \mathbf{c}_{t+h}$ denotes the phase velocity at time step $t+h$.
Second, we impose a temporal smoothness loss at the generation boundary:
\begin{equation}
\mathcal{L}_{\mathrm{smooth}}
=
\left\|
\hat{\mathbf{s}}_{t+1}
-
\mathbf{s}_{t}^{\mathrm{cond}}
\right\|_2^2.
\end{equation}
Third, we add a foot ground penetration penalty to discourage physically implausible foot placements. 
Let $\mathcal{F}$ denote the set of tracked foot points, and let $z_{t+h}^{(j)}$ be the vertical coordinate of foot point $j \in \mathcal{F}$ at time $t+h$. 
We penalize penetration below a small ground threshold $z_{\mathrm{g}}$ using
\begin{equation}
\mathcal{L}_{\mathrm{foot}}
=
\sum_{h=1}^{H}
\sum_{j\in\mathcal{F}}
\mathrm{ReLU}\!\left(z_{\mathrm{g}}-z_{t+h}^{(j)}\right),
\qquad
z_{\mathrm{g}}=0.035.
\end{equation}

The final training objective is
\begin{equation}
\mathcal{L}_{\mathrm{gen}}
=
\mathcal{L}_{\mathrm{VAE}}
+
\lambda_{1} \mathcal{L}_{\mathrm{phase}}
+
\lambda_{2} \mathcal{L}_{\mathrm{smooth}}
+
\lambda_{3} \mathcal{L}_{\mathrm{foot}}.
\end{equation}
Optimizing this objective yields a generative strike prior that not only reconstructs the captured motions, but also produces temporally structured and physically more plausible strike motions for downstream execution and policy learning.

\boldparagraph{Task noise and domain randomization}

Not all generated motions are suitable for downstream policy learning. 
Even when a synthesized motion is kinematically plausible, it may still be dynamically difficult to execute on the humanoid robot. 
We therefore introduce a tracker-based filtering stage to convert generated strike motions into physically executable motion references.

Given a generated motion sequence
\(
\hat{\tau}
=
\{\hat{\mathbf{s}}_1,\hat{\mathbf{s}}_2,\dots,\hat{\mathbf{s}}_T\},
\)
we feed it into a motion-tracking policy trained in simulation. 
The tracker attempts to follow the generated whole-body reference under robot dynamics and produces a physically realized rollout. 
We retain only those generated motions that can be stably tracked in simulation. 
This step removes samples that are plausible in motion space but incompatible with the robot's actuation limits or contact dynamics.

After tracker rollout, we further filter the retained motions using strike-event and phase consistency. 
First, we require the racket-ball contact event to lie near the temporal center of the motion clip, so that each motion contains both a sufficient preparation phase before contact and a meaningful recovery phase after contact. 
Second, we discard motions whose phase trajectory
\(
\{\hat{\mathbf{c}}_t\}_{t=1}^{T}
\)
deviates significantly from the expected strike-cycle pattern. 
These criteria remove motions with distorted timing, premature termination, or inconsistent strike ordering.

The final augmented strike library is constructed by merging the retained tracker-executable motions with the original motion-capture dataset. 
Compared with the raw mocap set, the resulting library provides substantially better coverage of strike points over the full table, while preserving the temporal structure and physical executability required for downstream whole-body policy learning.

\subsection{Policy: Task-oriented Motion Matching}
\label{sec:wbc_policy_learning}
Integrating large-scale primitive motions into task-oriented policy learning remains a key challenge at the intersection of imitation and reinforcement learning. Instead of adopting a hierarchical controller or distilling motion primitives into a latent skill model, we use a simple motion matching scheme over our enhanced strike motion library. Specifically, we retrieve the reference motion by nearest-neighbor search in a task-conditioned feature space defined by the relative strike target.
\begin{equation}
\begin{aligned}
\mathbf{p}_{\mathrm{hit}}^{\mathrm{rel}} &= \mathbf{p}_{\mathrm{hit}} + \boldsymbol{\epsilon} - \mathbf{p}_{\mathrm{anchor}}, \\
i^* &= \arg\min_{i\in\mathcal{M}} \left\| \mathbf{p}_{\mathrm{hit}}^{\mathrm{rel}} - \mathbf{p}_{\mathrm{target}}^{i} \right\|_2,
\end{aligned}
\end{equation}
where $\boldsymbol{\epsilon}$ is a small random perturbation that simulates target prediction errors during real-world deployment, $\mathbf{p}_{\mathrm{anchor}}$ is the anchor position, and $\mathbf{p}_{\mathrm{target}}^{i}$ is the strike target associated with the $i$-th motion. The matched motion is subsequently provided to the policy as part of the motion command observations, so that the current strike behavior is optimized toward the task objective while remaining consistent with the selected motion prior.

Our whole-body controller is trained with PPO under an asymmetric actor--critic framework. Importantly, the policy is not designed as a pure motion-tracking controller. Instead, it is optimized to satisfy task-level table-tennis objectives while being regularized by a matched reference motion prior. This formulation encourages the policy to produce task-effective yet natural whole-body behaviors, rather than merely replaying reference clips.

To support this objective, the policy observations are organized into three groups: \emph{motion commands}, which encode the selected reference motion and its phase-related cues; \emph{task commands}, which specify the desired strike target, strike velocity, and timing; and \emph{proprioceptive observations}, which describe the current robot state for closed-loop control. Following the asymmetric learning paradigm, the actor receives noisy and history-stacked observations to improve robustness, whereas the critic has access to cleaner and more privileged information for value estimation. The observation organization is summarized in Table~\ref{tab:wbc_asym_obs}.

\begin{table}[t]
\caption{Asymmetric observation design for actor and critic.}
\centering
\renewcommand{\arraystretch}{1.12}
\begin{adjustbox}{max width=\columnwidth,max totalheight=0.5\textheight,keepaspectratio}
\begin{tabular}{lcc}
\toprule
\textbf{Observation} & \textbf{Actor} & \textbf{Critic} \\
\midrule
\multicolumn{3}{l}{\textbf{\textit{Task Targets}}} \\
Strike time ($\tau_t$; noisy / clean) & \checkmark & \checkmark \\
Racket Target Vel ($\mathbf{v}^{\mathrm{hit}}_t$; noisy / clean) & \checkmark & \checkmark \\
Racket Target Pos ($\mathbf{p}^{\mathrm{hit}}_t$; noisy / clean) & \checkmark & \checkmark \\
Current time step & -- & \checkmark \\
Racket position & -- & \checkmark \\
Racket velocity & -- & \checkmark \\
Racket orientation error & -- & \checkmark \\

\midrule
\multicolumn{3}{l}{\textbf{\textit{Motion Targets}}} \\
Motion anchor position ($\mathbf{p}^{\mathrm{anchor}}$; noisy / clean) & \checkmark & \checkmark \\
Robot anchor position & \checkmark & \checkmark \\
Robot anchor orientation ($\mathbf{R}^{\mathrm{anchor}}$; noisy / clean) & \checkmark & \checkmark \\
Motion dof pos/vel & \checkmark & \checkmark \\
Motion body pos/orienation & -- & \checkmark \\

\midrule
\multicolumn{3}{l}{\textbf{\textit{Proprioception}}} \\
Base angular velocity ($\boldsymbol{\omega}_b$; noisy / clean) & \checkmark & \checkmark \\
Base linear velocity & -- & \checkmark \\
Joint positions ($\mathbf{q}$; noisy / clean) & \checkmark & \checkmark \\
Joint velocities ($\dot{\mathbf{q}}$; noisy / clean) & \checkmark & \checkmark \\
Previous action ($\mathbf{a}_{t-1}$) & \checkmark & \checkmark \\
\bottomrule
\end{tabular}
\end{adjustbox}
\label{tab:wbc_asym_obs}
\end{table}

Given these observations, the controller is trained with rewards that balance task execution, motion consistency, and control regularity. Among them, the task reward is the primary driver of ball-interaction quality, while the motion and regularization terms serve to preserve coordinated whole-body behavior and stable control.

We first detail the task reward. Since successful striking is governed primarily by the racket state near impact, we supervise the racket position and velocity errors only around the strike event, rather than enforcing rigid tracking throughout the motion. In addition, we define an orientation error to encourage alignment between the racket face normal and the desired strike velocity direction:
\begin{equation}
e_{\mathrm{ori}}(t)=
\arccos\!\left(
\left|
\mathbf{n}^{\mathrm{racket}}_t \cdot
\hat{\mathbf{v}}^{\mathrm{hit}}_t
\right|
\right),
\end{equation}
where $\mathbf{n}^{\mathrm{racket}}_t$ denotes the racket-face normal in the world frame and $\hat{\mathbf{v}}^{\mathrm{hit}}_t$ is the normalized desired strike velocity. The absolute value makes the reward symmetric with respect to the two opposite face directions, allowing both forehand and backhand strikes without imposing an explicit hand-specific constraint. We do not provide this orientation error as an explicit actor observation; instead, it is implicitly determined by the target strike velocity and the current robot state, and is learned through the motion-constrained policy optimization.

To concentrate supervision around the decisive contact phase without imposing rigid tracking throughout the motion, each term is shaped by a gated exponential reward:
\begin{equation}
r_j(t)=
\exp\!\left(-\frac{e_j(t)}{\sigma_j(t)}\right)\,
\mathbb{I}\!\left[\tau_t\in\mathcal{W}_j\right],
\qquad
j\in\{\mathrm{pos},\mathrm{ori},\mathrm{vel}\},
\end{equation}
where $\tau_t$ denotes the time-to-strike and $\mathcal{W}_j$ is the active supervision window. We use a tighter window for positional accuracy ($0.02\,\mathrm{s}$) and slightly wider windows ($0.1\,\mathrm{s}$) for orientation and velocity. This design enforces precise contact location near impact, while allowing a broader temporal margin for swing direction and speed, which we find helpful for avoiding overly sharp wrist accelerations and reducing the sim-to-real gap in task tracking.

The final task reward is defined as
\begin{equation}
r_t^{\mathrm{task}}
=
w_{\mathrm{pos}} r_{\mathrm{pos}}(t)
+
w_{\mathrm{ori}} r_{\mathrm{ori}}(t)
+
w_{\mathrm{vel}} r_{\mathrm{vel}}(t)
+
r_t^{\mathrm{succ}},
\end{equation}
where $r_t^{\mathrm{succ}}$ is a sparse success bonus activated when all strike constraints are simultaneously satisfied within the impact window.

Beyond task execution, the motion and regularization rewards serve to preserve coordinated whole-body behavior and stable control. Importantly, the wrist joints are excluded from the motion-tracking terms, preventing the reference motion from over-constraining the racket during ball interaction. This design decouples fine-grained strike execution from global motion regularization, allowing the wrist to adapt to task demands while the rest of the body remains consistent with the whole-body motion prior.

\boldparagraph{Task noise and domain randomization}
To improve robustness under dynamic interaction and mimic the gradual reduction of planning uncertainty near impact, we inject phase-dependent noise into the task commands. Denoting a generic task command by $\mathbf{c}_t$, we use
\begin{equation}
\tilde{\mathbf{c}}_t=\mathbf{c}_t+\boldsymbol{\xi}_t,\qquad
\|\boldsymbol{\xi}_t\|\propto s(\tau_t),
\end{equation}
where $s(\tau_t)$ is a decreasing function of the time-to-strike $\tau_t$, and $\boldsymbol{\xi}_t \sim \mathcal{N}(0, \sigma^2(\tau_t))$ is a Gaussian noise term whose scale is determined by $s(\tau_t)$. As a result, task perturbations are larger when the strike is still far away and gradually diminish near contact, reflecting the practical fact that target estimation and strike planning become more accurate as the ball approaches the hitting moment.

In addition, we inject random IMU noise into proprioceptive observations and apply standard domain randomization, including external push perturbations and friction randomization. These perturbations improve the policy's robustness to sensing uncertainty and sim-to-real discrepancies.

\boldparagraph{Adaptive Training Techniques}
To improve both workspace coverage and strike precision during training, we adopt an adaptive learning strategy with two components: \emph{adaptive region sampling} and \emph{adaptive task tracking}. The former reallocates training samples toward underperforming regions of the strike workspace, while the latter progressively sharpens the task-tracking reward as the policy improves.

For adaptive region sampling, we partition the target workspace into predefined regions and maintain a rolling success rate for each region based on recent trials. During training, the region with the lowest success rate is sampled with a higher probability, while the remaining regions are sampled uniformly. A strike target is then drawn within the selected region bounds. This simple strategy continuously reallocates training toward underperforming workspace zones, while still preserving exploration over the full strike space.

In addition, we employ adaptive task tracking by updating the reward scale \(\sigma_j\) from running tracking errors. As tracking improves, \(\sigma_j\) becomes smaller, yielding sharper reward shaping and progressively stricter strike constraints. This allows the policy to first acquire coarse task feasibility and then refine contact precision as learning proceeds.

\section{Egocentric Onboard Perception System}
\label{sec:system}

\begin{figure*}[t!]
  \centering
  \includegraphics[width= \textwidth]{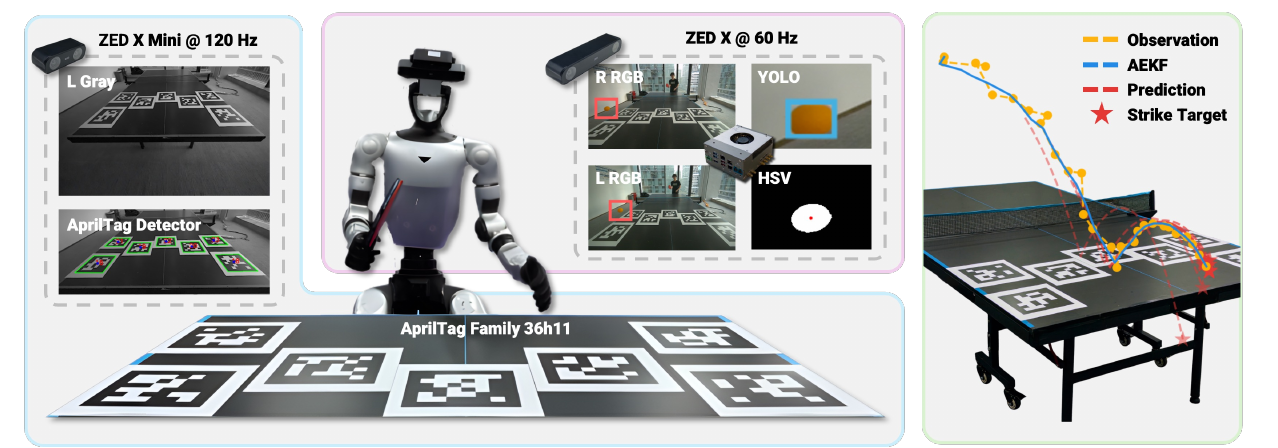} 
\caption{\textbf{Egocentric onboard perception system.} Our perception pipeline combines YOLO-based ball detection, AprilTag-based robot localization, and adaptive Kalman filtering for state estimation. The resulting system provides robust real-time ball trajectory prediction and strike-target estimation using only onboard sensing.}
  \label{fig:onboard}
\end{figure*}

We present an egocentric onboard perception system (\cref{fig:onboard}) that enables the first outdoor consecutive humanoid table-tennis rallies. 
Closing the perception–action loop under highly dynamic conditions is inherently challenging. To address this, we adopt a dual-modality perception framework (\cref{fig:method}) with two complementary operating modes. 
A MoCap-based pipeline provides high-accuracy, high-frequency sensing for rapid validation of the full perception–control stack in laboratory settings, while the same pipeline is deployed with ego-view onboard sensing for real-world operation.

In both modes, the perception system estimates two key quantities: the ping-pong ball position in the table frame $\mathbf{p}^{\mathcal{T}}_{\mathrm{ball}}$, and the robot torso pose in the origin frame $\mathcal{O}$, represented by ${}^{\mathcal{O}}T_{\mathrm{Torso}}$, where $\mathcal{O}$ denotes the reference world frame used during policy training.

\paragraph{Motion Capture (MoCap) mode}
For controlled experiments, a professional optical MoCap system operates at 120\,Hz. The ball is coated with retro-reflective markers and tracked directly from the MoCap stream. The robot torso pose is obtained via a rigid-body tracker mounted on the torso, which is registered within the system to localize the robot in the origin frame.

\paragraph{Ego-view camera mode}  
For in-the-wild deployment, a dual-camera setup is used. A head-mounted ZED\,X performs long-range ball detection at 60\,Hz, while a downward-tilted ZED\,X Mini provides robot localization at 120\,Hz. 
The ball position is obtained via stereo triangulation in the camera frame, while the table pose is estimated via AprilTag detection. Through pre-calibrated extrinsic transformations, the ball position in $\mathcal{T}$ and the torso pose in $\mathcal{O}$ are computed, where $\mathcal{O}$ is rigidly defined relative to $\mathcal{T}$.

Both modalities share an identical processing pipeline —comprising an Adaptive Extended Kalman Filter (AEKF), a physics-based trajectory predictor, and a racket strike planner—and differ only in modality-specific hyperparameters. This unified design ensures consistent state estimation across sensing modes and allows MoCap to serve as a reference for tuning the ego-view system.

\subsection{Coordinate Frame Definitions}
\begin{table}[t!]
  \centering
  \footnotesize
  \renewcommand{\arraystretch}{1.15}
  
  \caption{Reference frames used in the perception system.}
  \label{tab:frames}
  
  \begin{tabularx}{\columnwidth}{@{}l l >{\raggedright\arraybackslash}X@{}}
    \toprule
    \textbf{Frame} 
      & \textbf{Symbol} 
      & \textbf{Description} \\
    \midrule
    
    World (MoCap) 
      & $\mathcal{W}$ 
      & Global MoCap coordinate frame \\
    
    Table 
      & $\mathcal{T}$ 
      & Origin at table centre; $x$ toward opponent, $z$ upward \\
    
    Origin 
      & $\mathcal{O}$ 
      & On the floor, with axes aligned with $\mathcal{T}$; 
        the offset from the robot-side table edge is a fixed scalar 
        $d_{\mathrm{orig}}$ set per deployment or task. \\
    
    Camera (ZED\,X) 
      & $\mathcal{C}_1$ 
      & Left-eye frame of the head-mounted ZED\,X stereo pair 
        (ball triangulation) \\
    
    Camera (ZED\,X Mini) 
      & $\mathcal{C}_2$ 
      & Left-eye frame of the ZED\,X Mini 
        (AprilTag-based ego-localization) \\
    
    \bottomrule
  \end{tabularx}
\end{table}

We define five reference frames used throughout the system (Table~\ref{tab:frames}). The rigid transformations between frames are pre-calibrated:
\begin{equation}
  {}^{\mathcal{W}}T_{\mathcal{T}}, \quad
  {}^{\mathcal{W}}T_{\mathcal{O}}, \quad
  {}^{\mathcal{C}_2}T_{\mathcal{C}_1}, \quad
  {}^{\mathrm{Torso}}T_{\mathcal{C}_2}.
\end{equation}

All transformations are represented as $4\times4$ homogeneous matrices.
The table and origin frames are fixed relative to each other:
\[
{}^{\mathcal{O}}T_{\mathcal{T}} =
  {}^{\mathcal{W}}T_{\mathcal{O}}^{-1} \cdot {}^{\mathcal{W}}T_{\mathcal{T}},
\]
and are calibrated once prior to each session.

\subsection{MoCap-Based Sensing Pipeline}

\boldparagraph{Ball localization}
The ball is uniformly coated with retroreflective material, allowing the MoCap system to treat it as a single marker. The system reports its 3D position $\mathbf{p}^{\mathcal{W}}_{\mathrm{ball}}$ at 120\,Hz, which is transformed into the table frame as:
\begin{equation}
  \mathbf{p}^{\mathcal{T}}_{\mathrm{ball}}
    = {}^{\mathcal{W}}T_{\mathcal{T}}^{-1} \cdot
      \tilde{\mathbf{p}}^{\mathcal{W}}_{\mathrm{ball}}.
\end{equation}

\boldparagraph{Robot self-localization}
A rigid-body tracker mounted on the robot provides ${}^{\mathcal{W}}T_{\mathrm{Tracker}}$. With the pre-calibrated extrinsic ${}^{\mathrm{Tracker}}T_{\mathrm{Torso}}$, the torso pose in $\mathcal{O}$ is computed as:
\begin{equation}
  {}^{\mathcal{O}}T_{\mathrm{Torso}}
    = {}^{\mathcal{W}}T_{\mathcal{O}}^{-1} \cdot
      {}^{\mathcal{W}}T_{\mathrm{Tracker}} \cdot
      {}^{\mathrm{Tracker}}T_{\mathrm{Torso}}^{-1}.
\end{equation}
The result is streamed to the low-level controller at 120\,Hz.

\subsection{Ego-View Camera Sensing Pipeline}

\boldparagraph{Hardware}
A ZED\,X stereo camera mounted on the robot head detects the ball at 60\,Hz via stereo triangulation. A ZED\,X Mini, mounted with a downward tilt, provides ego-localization at 120\,Hz using AprilTag detection from its left camera stream.

\boldparagraph{Ball Detection and 3-D Localization}
Ball detection is implemented as a three-stage pipeline on a Jetson Orin:

\begin{enumerate}
  \item \textit{Coarse detection via YOLO.}  
    A YOLO detector trained on standard yellow balls predicts bounding boxes in the left and right RGB images, providing a reliable spatial prior.

  \item \textit{Sub-pixel refinement via HSV segmentation.}  
  Within each bounding box, HSV-based color segmentation isolates the ball region, offering improved robustness to illumination changes. The centroid of the resulting mask defines the ball center $(u, v)$.

  \item \textit{Stereo triangulation.}  
  Corresponding pixel coordinates $(u_L, u_R)$ are used to triangulate the 3D position $\mathbf{p}^{\mathcal{C}_1}_{\mathrm{ball}}$ in the camera frame.
\end{enumerate}

\boldparagraph{Robot Self-Localisation via AprilTag}
AprilTag-based localization is performed using the ZED\,X Mini's left camera. Seven tags with known positions $\{{}^{\mathcal{T}}\mathbf{P}_{i}^{j}\}$ are placed on the table. Detected tag corners are aggregated into a unified Perspective-n-Point (PnP) problem, and ${}^{\mathcal{C}_2}T_{\mathcal{T}}$ is estimated using RANSAC-PnP.

To maintain 120\,Hz performance, detection is restricted to a dynamically updated region of interest (ROI), defined by projecting table corners using the previous pose estimate.

\boldparagraph{Torso pose derivation}
Given ${}^{\mathcal{T}}T_{\mathcal{C}_2}$ and ${}^{\mathrm{Torso}}T_{\mathcal{C}_2}$:
\begin{equation}
  {}^{\mathcal{O}}T_{\mathrm{Torso}}
    = {}^{\mathcal{O}}T_{\mathcal{T}} \cdot
      {}^{\mathcal{T}}T_{\mathcal{C}_2} \cdot
      \bigl({}^{\mathrm{Torso}}T_{\mathcal{C}_2}\bigr)^{-1}.
\end{equation}

\boldparagraph{Ball position in the table frame}
\begin{equation}
  {}^{\mathcal{T}}\mathbf{p}_{\mathrm{ball}}
    = {}^{\mathcal{T}}T_{\mathcal{C}_2} \cdot
      {}^{\mathcal{C}_2}T_{\mathcal{C}_1} \cdot
      {}^{\mathcal{C}_1}\tilde{\mathbf{p}}_{\mathrm{ball}}.
\end{equation}
\subsection{Ball State Estimation: Adaptive Extended Kalman Filter}

Raw ball position measurements are noisy and exhibit discontinuities at
table bounces.  We design an Adaptive Extended Kalman Filter (AEKF) that fuses position observations with a physics-based motion
model to jointly estimate ball position and velocity, while robustly handling bounce events. The process is illustrated in \Cref{alg:aekf}.

\boldparagraph{State Space} The filter state stacks position and velocity in Table coordinates,
\begin{equation}
  \mathbf{x}
    = \bigl[p_x,\; p_y,\; p_z,\; v_x,\; v_y,\; v_z\bigr]^{\top}
    \in \mathbb{R}^{6}.
\end{equation}

The observation model is position-only,
\begin{equation}
\begin{aligned}
     \mathbf{z}_k &= \mathbf{H}\,\mathbf{x}_k + \mathbf{n}_k, \quad \mathbf{n}_k \sim \mathcal{N}(\mathbf{0},\,\mathbf{R}(d_k)). \\
     \mathbf{H} &= \bigl[\,\mathbf{I}_{3} \;\big|\; \mathbf{0}_{3\times 3}\,\bigr] \in \mathbb{R}^{3\times 6}.
\end{aligned}
\end{equation}
Here $\mathbf{0}_{3\times 3}$ denotes the $3\times 3$ zero matrix; $\mathbf{R}(d_k)$ is specified below.

\boldparagraph{Physics-Based Prediction Model} The ball's flight dynamics follow a quadratic aerodynamic drag model:
\begin{equation}
  \mathbf{a}
    = -k\,\|\mathbf{v}\|\,\mathbf{v} - g\,\hat{\mathbf{z}}.
\end{equation}
where $k$ is the drag coefficient, $\hat{\mathbf{z}}$ is the upward vertical unit vector, and $g$ is the gravitational acceleration magnitude.
The state $\mathbf{x}=[\mathbf{p}^{\top},\mathbf{v}^{\top}]^{\top}$ is advanced over $\Delta t$ by

\begin{equation}
f_{\text{drag}}: \left\{
\begin{aligned}
  \mathbf{p}_{k} &= \mathbf{p}_{k-1} + \mathbf{v}_{k-1}\,\Delta t + \tfrac{1}{2}\,\mathbf{a}(\mathbf{v}_{k-1})\,\Delta t^{2}, \\
  \mathbf{v}_{k} &= \mathbf{v}_{k-1} + \mathbf{a}(\mathbf{v}_{k-1})\,\Delta t.
\end{aligned}
\right.
\end{equation}

After the free-flight update, a table collision is applied when the ball lies inside the table rectangle
$|p_x|<L_x$, $|p_y|<L_y$, is at or below the collision height $p_z \le z_c$, and is moving downward ($v_z<0$).
The post-impact state uses horizontal restitution $C_h$ and vertical restitution $C_v$:
\begin{equation}
  \mathbf{v} \leftarrow \mathrm{diag}(C_h,\,C_h,\,-C_v)\,\mathbf{v}, \quad
  p_z \leftarrow 2 z_c - p_z.
\end{equation}

The process noise
$\mathbf{Q}_k
  = \mathrm{diag}(q_{\mathrm{pos}},q_{\mathrm{pos}},q_{\mathrm{pos}},
                  q_{\mathrm{vel}},q_{\mathrm{vel}},q_{\mathrm{vel}})$
is normalised by the nominal time step $\Delta t_0 = 1/f_s$:
\begin{equation}
  q_{\mathrm{pos}}
    = Q_{\mathrm{pos}}^{\mathrm{base}}
      \left(\frac{\Delta t}{\Delta t_0}\right)^{\!2},
  \quad
  q_{\mathrm{vel}}
    = Q_{\mathrm{vel}}^{\mathrm{base}}
      \frac{\Delta t}{\Delta t_0}.
\end{equation}
where $1/f_s$ is the capture frequency of the camera

\boldparagraph{Adaptive Observation Noise} Let $d_k$ denote the camera--ball distance at time $k$.
The measurement noise covariance is isotropic and grows with $d_k$,
\begin{equation}
  \mathbf{R}(d_k)
    = r(d_k)\,\mathbf{I}_{3},
  \quad
  r(d) = R_{\mathrm{base}}\,\bigl(1 + \beta\,d\bigr).
\end{equation}
where $R_{\mathrm{base}}$ is the baseline variance (in $\mathrm{m}^2$) on each position axis and $\beta>0$ is a tunable distance-sensitivity gain.
Closer range yields smaller $r(d)$ and tighter fusion of $\mathbf{z}_k$; larger $d$ inflates $\mathbf{R}$ to reflect weaker confidence in triangulation.

\begin{algorithm}[t]
\caption{Adaptive Extended Kalman Filter}
\label{alg:aekf}
\If{not initialised}{
  $\mathbf{x} \leftarrow [\mathbf{z}_k^{\top},\mathbf{v}_{\mathrm{init}}^{\top}]^{\top}$,\ $\mathbf{P} \leftarrow \mathbf{P}_{\mathrm{init}}$,\ $t_{\mathrm{last}} \leftarrow t_k$\;
  \KwRet\;
}
$\Delta t \leftarrow t_k - t_{\mathrm{last}}$\;
\If{$\Delta t > \Delta t_{\max}$}{re-init from $(\mathbf{z}_k,t_k)$\;\KwRet\;}
read $(\mathbf{p},\mathbf{v})$ from $\mathbf{x}$\;
$\mathbf{F} \leftarrow \partial f_{\mathrm{drag}}/\partial \mathbf{x}$ at this $(\mathbf{p},\mathbf{v})$\;
$\mathbf{a} \leftarrow -k\,\|\mathbf{v}\|\,\mathbf{v} - g\,\hat{\mathbf{z}}$\;
$\mathbf{p} \leftarrow \mathbf{p} + \mathbf{v}\,\Delta t + \tfrac{1}{2}\,\mathbf{a}\,\Delta t^{2}$;\ $\mathbf{v} \leftarrow \mathbf{v} + \mathbf{a}\,\Delta t$\;
\If{$|p_x|<L_x$ \textbf{and} $|p_y|<L_y$ \textbf{and} $p_z \le z_c$ \textbf{and} $v_z<0$}{
  $v_x \leftarrow C_h v_x$,\ $v_y \leftarrow C_h v_y$,\ $v_z \leftarrow -C_v v_z$,\ $p_z \leftarrow 2 z_c - p_z$\;
}
write $(\mathbf{p},\mathbf{v})$ to $\mathbf{x}$\;
$q_{\mathrm{pos}} \leftarrow Q_{\mathrm{pos}}^{\mathrm{base}}(\Delta t/\Delta t_0)^2$,\ $q_{\mathrm{vel}} \leftarrow Q_{\mathrm{vel}}^{\mathrm{base}}(\Delta t/\Delta t_0)$\;
$\mathbf{Q} \leftarrow \mathrm{diag}(q_{\mathrm{pos}},q_{\mathrm{pos}},q_{\mathrm{pos}},q_{\mathrm{vel}},q_{\mathrm{vel}},q_{\mathrm{vel}})$\;
$\mathbf{P} \leftarrow \mathbf{F}\mathbf{P}\mathbf{F}^{\top}+\mathbf{Q}$\;
\If{$\hat{p}_x-z_{k,x}>\tau_x$ \textbf{and} $\hat{v}_x>0$}{re-init from $(\mathbf{z}_k,t_k)$\;\KwRet\;}
$\mathbf{R}(d_k) = r(d_k)\,\mathbf{I}_{3}, \quad
  r(d) = R_{\mathrm{base}}\,\bigl(1 + \beta\,d\bigr)$\;
standard Kalman measurement update on $(\mathbf{x},\mathbf{P})$\;
$t_{\mathrm{last}} \leftarrow t_k$\;
\KwRet $(\mathbf{p},\mathbf{v})$\;
\end{algorithm}

\boldparagraph{Initialisation and Resets}
At initialization, the position is set to the first measurement $\mathbf{z}_0$, while the velocity is initialized with a fixed prior. The initial covariance $\mathbf{P}_{0|0}$ is chosen as a diagonal matrix with large entries. The filter is reset using the current measurement if no observation is received for longer than $\Delta t_{\max}$, or if a return is detected, indicated by the observed $p_x$ falling below the predicted value by more than $\tau_x$.

\subsection{Ball Trajectory Prediction}

Given the current filtered ball state in table coordinates,
$(\hat{\mathbf{p}}^{t},\,\hat{\mathbf{v}}^{t})$,
the future trajectory is propagated using the same flight model as in state estimation,
including quadratic air drag, gravity, and table impacts with restitution coefficients $(C_h, C_v)$, using a fixed small time step.

Candidate arrival times $\tau$ are determined in two stages.
\begin{enumerate}
\item \boldparagraph{Initial search}
When the ball is approaching (e.g., based on along-table velocity and position gating), candidate times are sampled over
$[T_{\det}^{\min},\,T_{\det}^{\max}]$ with a coarse step (e.g., $0.01\,\mathrm{s}$).
For each $\tau$, the state is propagated in $\mathcal{T}$ and mapped to the origin frame $\mathcal{O}$. Samples are retained only if the predicted position lies within the admissible strike volume
\begin{equation}
  \begin{aligned}
    \mathcal{S} = \bigl\{(x,y,z) :\;
      & x \in [x_{\min},\,x_{\max}],\\
      & y \in [y_{\min},\,y_{\max}], \\
      & z \in [z_{\min},\,z_{\max}]
    \bigr\}\;\text{(m, in }\mathcal{O}\text{)}.
  \end{aligned}
\end{equation}
Among feasible candidates, $\tau$ is selected to minimise
$\bigl|\hat{p}_x^{(o)}(\tau) - c_{s,x}\bigr|$, where $c_{s,x}$ denotes the strike-plane location in $\mathcal{O}$.

\item \boldparagraph{Online refinement}
Once a strike phase is active, $\tau$ is refined over a local window
$\tau \in [\tau_{\mathrm{cur}} - W_\tau,\, \tau_{\mathrm{cur}} + W_\tau]$ with step $\Delta\tau$,
while $\tau>0$ and the ball remains on the incoming side of the strike plane ($p_x^{(o)} > c_{s,x}$).
The minimiser of $\bigl|\hat{p}_x^{(o)}(\tau) - c_{s,x}\bigr|$ is selected without enforcing $\hat{\mathbf{p}}^{(o)}(\tau)\in\mathcal{S}$.
\end{enumerate}

The outputs are the time-to-strike $\tau$, and the predicted position and velocity at $\tau$ in $\mathcal{O}$, optionally clipped to $\mathcal{S}$ and velocity bounds prior to racket planning.
If no new observation is available, $\tau$ is decremented at the control rate and clamped to $[\tau_{\min},\,\tau_{\max}]$. When $\tau$ becomes negative, the strike phase terminates and the system waits for the next incoming ball.

\subsection{Racket Strike Planning}

Given the predicted impact position $\mathbf{p}_{\mathrm{hit}}$ and incoming velocity $\mathbf{v}_{\mathrm{in}}$, we compute the required racket velocity $\mathbf{V}_r$ and surface normal $\hat{n}$.

\boldparagraph{Desired outgoing velocity}
For a target landing position $\mathbf{p}_{\mathrm{target}}$ and flight time $T_f$, the required outgoing ball velocity $\mathbf{v}_{\mathrm{out}}$ is obtained analytically under a linearised drag model $\dot{\mathbf{v}} = -k\mathbf{v} + \mathbf{g}$:
\begin{align}
  v_{\mathrm{out},i}
    &= \frac{\bigl(p_{\mathrm{target},i}
               - p_{\mathrm{hit},i}\bigr)\,k}
             {1 - e^{-kT_f}},
  \quad i \in \{x, y\}, \\[6pt]
  v_{\mathrm{out},z}
    &= \frac{g_z}{k}
       + \frac{\bigl(p_{\mathrm{target},z} - p_{\mathrm{hit},z}\bigr)
               - g_z T_f / k}
              {\bigl(1 - e^{-kT_f}\bigr)/k}.
\end{align}

\boldparagraph{Racket motion via collision model}
Assuming the racket velocity is aligned with its surface normal, $\mathbf{V}_r = V_n\,\hat{n}$, a frictionless normal collision yields
\begin{equation}
  v_{\mathrm{out},n}
    = v_{\mathrm{in},n} - (1+e)\,(v_{\mathrm{in},n} - V_n),
\end{equation}
where $e$ is the restitution coefficient. Since the velocity change is along $\hat{n}$,
\begin{align}
  \hat{n}
    &= \frac{\mathbf{v}_{\mathrm{out}} - \mathbf{v}_{\mathrm{in}}}
            {\|\mathbf{v}_{\mathrm{out}} - \mathbf{v}_{\mathrm{in}}\|}, \\[6pt]
  V_n
    &= v_{\mathrm{in},n}
       - \frac{v_{\mathrm{in},n} - v_{\mathrm{out},n}}{1 + e}.
\end{align}
The resulting $\hat{n}$ and $\mathbf{V}_r = V_n\,\hat{n}$ are sent to the low-level controller in the $\mathcal{O}$ frame.

\section{Evaluations}
\label{sec:evaluations}
In this section, we provide a comprehensive evaluation of the proposed \methodname\ system from three perspectives. First, we examine its overall performance on humanoid table tennis and compare it with representative baselines from reinforcement learning, motion imitation, and prior system-level solutions. Second, we evaluate the proposed motion-VAE-based data augmentation scheme and analyze how the expanded strike dataset benefits policy learning. Third, we investigate the proposed perception system through extensive ablation studies on its key modules. The evaluation metrics are summarized in \cref{tab:metrics}, we choose the following methods as baselines:
\begin{itemize}[leftmargin=1.5em]
    \item \textit{PPO}~\cite{schulman2017proximalpolicyoptimizationalgorithms}: We directly train the policy using vanilla PPO with only task rewards, without introducing any motion prior. This baseline evaluates task performance without motion guidance.

    \item \textit{Mimic}: We adopt BeyondMimic ~\cite{Liao2025BeyondMimicFM} as a representative motion-imitation baseline. In this setting, we sample strike targets and matched motions, but do not introduce task rewards; the policy is trained only to track reference motions. This baseline evaluates the pure motion-imitation capability of the policy without task constraints.
    
    \item \textit{HITTER}~\cite{su2025hitterhumanoidtabletennis}: We adopt the training paradigm of HITTER, a state-of-the-art humanoid table-tennis system, where the robot imitates only forehand and backhand swing motions without lower-body tracking. This serves as a system-level baseline for comparison.
\end{itemize}

\begin{table}[t]
\caption{Evaluation metrics.}
\label{tab:metrics}
\centering
\footnotesize
\renewcommand{\arraystretch}{1.16}
\setlength{\tabcolsep}{4pt}
\resizebox{0.48\textwidth}{!}{%
\begin{tabular}{@{}>{\raggedright\arraybackslash}p{2.35cm}>{\raggedright\arraybackslash}p{6.05cm}@{}}
\toprule
\textbf{Metric} & \textbf{Definition} \\

$E_{\text{p}}$ 
& Racket position error, $\|\mathbf{p}^{\mathrm{racket}}-\mathbf{p}^{\mathrm{hit}}\|_2$ (cm). \\

$E_{\text{v}}$ 
& Racket velocity error, $\|\mathbf{v}^{\mathrm{racket}}-\mathbf{v}^{\mathrm{hit}}\|_2$. \\

$E_{\text{o}}$ 
& Racket orient error, $100 * \arccos\!\left(\left|\mathbf{n}^{\mathrm{racket}}\!\cdot\!\hat{\mathbf{v}}^{\mathrm{hit}}\right|\right)$. \\

$E_{\mathrm{predpos}}$ 
& Mean predicted strike position error (cm). \\

$E_{\mathrm{predvel}}$ 
& Mean predicted strike velocity error (m/s). \\

$SR$ 
& $E_{\text{p(cm)}}<0.04\,\mathrm{m}$, $E_{\text{o}}<0.05$, and $E_{\text{v}}<0.5$. \\

$SR_{\text{det}}$
& Successful detection of the incoming ball. \\

$SR_{\text{hit}}$
& Successful racket--ball contact. \\

$SR_{\text{return}}$
& Successful returns to the opponent side. \\

$E_{\mathrm{mpjpe}} \cite{He2024OmniH2OUA}$ 
& MPJPE with respect to the reference motion. \\

$E_{\mathrm{orientation}}$ 
& Body Orientation error to the reference motion. \\

$E_{\mathrm{smoothness}}$ 
& Mean joint acceleration per step, $\frac{1}{N}\sum_i |\ddot{q}_i|$. \\

$E_{\mathrm{torque}}$ 
& Mean applied torque per step, $\frac{1}{N}\sum_i |\tau_i|$. \\

$SR_{\text{balance}}$
& The robot maintains stable posture during the trail. \\

\bottomrule
\end{tabular}%
}
\end{table}

\subsection{Policy Performance Evaluation}
\begin{table*}[t!]
  \centering
  \caption{Policy evaluations in simulation.}
  \label{tab:Simulation Results}
  \resizebox{\textwidth}{!}{%
  \begin{tabular}{lcccccccccc}
    \toprule
    \multirow{2}{*}{\textbf{Metrics}} &
      \multicolumn{4}{c}{\textbf{Task Tracking}} &
      \multicolumn{2}{c}{\textbf{Motion Tracking}}&
      \multicolumn{2}{c}{\textbf{Regularization}}\\
      \cmidrule(lr){2-5}\cmidrule(lr){6-7}\cmidrule(lr){8-9}
     & $SR\!\uparrow$ & $E_{\text{p(cm)}}\!\downarrow$ & $E_{\text{v}}\!\downarrow$ &  $E_{\text{o}}\!\downarrow$ 
     &  $E_{\text{mpjpe}}\!\downarrow$ & $E_{\text{orientation}}\!\downarrow$
    &  $E_{\text{smoothness}}\!\downarrow$ & $E_{\text{torque}}\!\downarrow$\\     
    \midrule
    \rowcolor{gray!18}\multicolumn{8}{l}{\textbf{\textit{Baseline Comparisons}}} \\
    Mimic  & $N/A$ & $N/A$ & $N/A$ & $N/A$ & $\bm{38.23}\!\pm\! \mathsmaller{0.099}$ & $\bm{0.15}\!\pm\! \mathsmaller{0.000}$ & $\bm{16.60}\!\pm\! \mathsmaller{0.075}$ & $5.96\!\pm\! \mathsmaller{0.006}$ \\
    PPO \cite{schulman2017proximalpolicyoptimizationalgorithms} & $75.28\!\pm\! \mathsmaller{0.536}$ & $4.63\!\pm\! \mathsmaller{0.015}$ & $0.34\!\pm\! \mathsmaller{0.006}$ & $3.12\!\pm\! \mathsmaller{0.025}$ & $146.49\!\pm\! \mathsmaller{0.215}$ & $0.88\!\pm\! \mathsmaller{0.000}$ & $24.25\!\pm\! \mathsmaller{0.059}$ & $5.83\!\pm\! \mathsmaller{0.006}$ \\
    HITTER \cite{su2025hitterhumanoidtabletennis}    & $\bm{86.63}\!\pm\! \mathsmaller{0.260}$ & $4.69\!\pm\! \mathsmaller{0.017}$ & $\bm{0.33}\!\pm\! \mathsmaller{0.000}$ & $\bm{3.02}\!\pm\! \mathsmaller{0.015}$ &  $100.05\!\pm\! \mathsmaller{0.156}$ & $0.47\!\pm\! \mathsmaller{0.000}$ & $20.77\!\pm\! \mathsmaller{0.040}$ & $5.79\pm\! \mathsmaller{0.000}$\\
    \methodname (Ours) & $86.38\!\pm\! \mathsmaller{0.078}$ & $\bm{4.42}\!\pm\! \mathsmaller{0.029}$ & $0.35\!\pm\! \mathsmaller{0.000}$ & $4.17\!\pm\! \mathsmaller{0.118}$ & $75.01\!\pm\! \mathsmaller{0.038}$ & $0.32\!\pm\! \mathsmaller{0.006}$ & $20.99\!\pm\! \mathsmaller{0.010}$ & $\bm{5.58}\!\pm\! \mathsmaller{0.000}$\\
    \midrule
    \rowcolor{gray!18}\multicolumn{8}{l}{\textbf{\textit{Ablation Data Scale}}} \\
    64 mocap clips     & $73.61\!\pm\! \mathsmaller{0.602}$ & $4.97\!\pm\! \mathsmaller{0.055}$ & $0.46\!\pm\! \mathsmaller{0.006}$ & $7.77\!\pm\! \mathsmaller{0.050}$ & $84.54\!\pm\! \mathsmaller{0.161}$ & $0.35\!\pm\! \mathsmaller{0.000}$ & $20.83\!\pm\! \mathsmaller{0.021}$ & $5.38\!\pm\! \mathsmaller{0.010}$\\
    200 mocap clips     & $76.86\!\pm\! \mathsmaller{0.140}$ & $4.98\!\pm\! \mathsmaller{0.110}$ & $0.39\!\pm\! \mathsmaller{0.000}$ & $4.50\!\pm\! \mathsmaller{0.072}$ & $81.90\!\pm\! \mathsmaller{0.025}$ & $0.35\!\pm\! \mathsmaller{0.000}$ & $21.01\!\pm\! \mathsmaller{0.035}$ & $5.56\!\pm\! \mathsmaller{0.010}$\\
    400 mocap clips    & $76.55\!\pm\! \mathsmaller{0.191}$ & $4.83\!\pm\! \mathsmaller{0.055}$ & $0.39\!\pm\! \mathsmaller{0.000}$ & $4.77\!\pm\! \mathsmaller{0.271}$ & $84.00\!\pm\! \mathsmaller{0.086}$ & $0.35\!\pm\! \mathsmaller{0.000}$ & $20.57\!\pm\! \mathsmaller{0.012}$ & $5.42\!\pm\! \mathsmaller{0.006}$\\
    400-2k-gen   & $82.32\!\pm\! \mathsmaller{0.471}$ & $4.49\!\pm\! \mathsmaller{0.032}$ & $0.39\!\pm\! \mathsmaller{0.006}$ & $3.96\!\pm\! \mathsmaller{0.067}$ & $74.57\!\pm\! \mathsmaller{0.283}$ & $0.32\!\pm\! \mathsmaller{0.000}$ & $21.33\!\pm\! \mathsmaller{0.045}$ & $5.62\!\pm\! \mathsmaller{0.010}$\\
    60-5k-gen             & $83.06\!\pm\! \mathsmaller{0.594}$ & $4.47\!\pm\! \mathsmaller{0.066}$ & $0.36\!\pm\! \mathsmaller{0.000}$ & $4.44\!\pm\! \mathsmaller{0.165}$ & $\bm{68.70}\!\pm\! \mathsmaller{0.075}$ & $\bm{0.30}\!\pm\! \mathsmaller{0.006}$ & $21.27\!\pm\! \mathsmaller{0.055}$ & $5.67\!\pm\! \mathsmaller{0.006}$\\
    200-5k-gen           & $85.29\!\pm\! \mathsmaller{0.070}$ & $\bm{4.20}\!\pm\! \mathsmaller{0.026}$ & $\bm{0.34}\!\pm\! \mathsmaller{0.000}$ & $4.02\!\pm\! \mathsmaller{0.132}$ & $76.52\!\pm\! \mathsmaller{0.122}$ & $0.33\!\pm\! \mathsmaller{0.006}$ & $21.54\!\pm\! \mathsmaller{0.015}$ & $5.60\!\pm\! \mathsmaller{0.000}$\\
    400-5k-gen (\methodname) & $\bm{86.38}\!\pm\! \mathsmaller{0.078}$ & $4.42\!\pm\! \mathsmaller{0.029}$ & $0.35\!\pm\! \mathsmaller{0.000}$ & $4.17\!\pm\! \mathsmaller{0.118}$ & $75.01\!\pm\! \mathsmaller{0.038}$ & $0.32\!\pm\! \mathsmaller{0.006}$ & $\bm{20.99}\!\pm\! \mathsmaller{0.010}$ & $5.58\!\pm\! \mathsmaller{0.000}$\\
    400-5k-gen (w/o tracker)  & $85.53\!\pm\! \mathsmaller{0.379}$ & $4.51\!\pm\! \mathsmaller{0.135}$ & $\bm{0.34}\!\pm\! \mathsmaller{0.000}$ & $\bm{3.70}\!\pm\! \mathsmaller{0.130}$ & $91.02\!\pm\! \mathsmaller{0.275}$ & $0.37\!\pm\! \mathsmaller{0.006}$ & $21.05\!\pm\! \mathsmaller{0.058}$ & $\bm{5.29}\!\pm\! \mathsmaller{0.006}$\\
    400-10k-gen   & $85.88\!\pm\! \mathsmaller{0.199}$ & $4.35\!\pm\! \mathsmaller{0.038}$ & $0.36\!\pm\! \mathsmaller{0.000}$ & $4.25\!\pm\! \mathsmaller{0.162}$ & $76.21\!\pm\! \mathsmaller{0.178}$ & $0.32\!\pm\! \mathsmaller{0.000}$ & $21.32\!\pm\! \mathsmaller{0.021}$ & $5.63\!\pm\! \mathsmaller{0.012}$\\
    \midrule
    \rowcolor{gray!18}\multicolumn{8}{l}{\textbf{\textit{Ablation training techniques}}} \\
    w/o region adaptive sampling & $82.72\!\pm\! \mathsmaller{0.321}$ & $4.70\!\pm\! \mathsmaller{0.055}$ & $0.35\!\pm\! \mathsmaller{0.000}$ & $4.08\!\pm\! \mathsmaller{0.096}$ & $74.22\!\pm\! \mathsmaller{0.065}$ & $0.31\!\pm\! \mathsmaller{0.006}$ & $20.87\!\pm\! \mathsmaller{0.035}$ & $5.70\!\pm\! \mathsmaller{0.000}$ \\
    w/o adaptive tracking sigma & $22.60\!\pm\! \mathsmaller{0.800}$ & $11.94\!\pm\! \mathsmaller{0.222}$ & $0.62\!\pm\! \mathsmaller{0.006}$ & $35.49\!\pm\! \mathsmaller{0.410}$ & $62.21\!\pm\! \mathsmaller{0.176}$ & $0.25\!\pm\! \mathsmaller{0.000}$ & $18.73\!\pm\! \mathsmaller{0.025}$ & $5.44\!\pm\! \mathsmaller{0.010}$  \\ \bottomrule
  \end{tabular}}
\end{table*}

\begin{figure*}[t!]
  \centering
  \includegraphics[width= 1.05\textwidth]{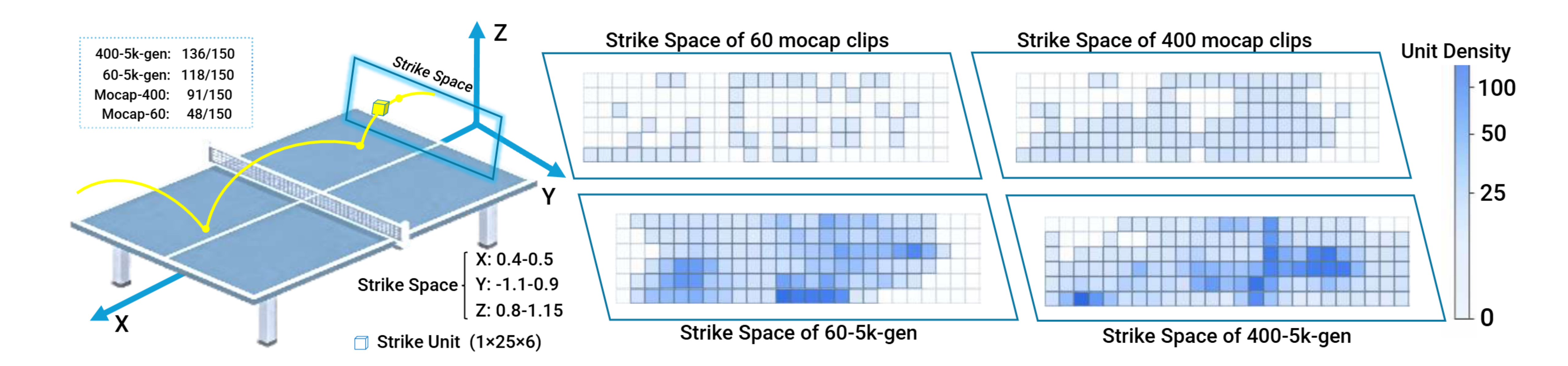} 
\caption{\textbf{Motion-VAE expands strike-space coverage.} Compared with the original mocap dataset, Motion-VAE-generated motions produce a much broader distribution of strike targets over the reachable hitting workspace, substantially improving data coverage for downstream policy learning.}
  \label{fig:mvae_generalization}
\end{figure*}

\begin{figure}[t!]
  \centering
  \includegraphics[width= 0.47\textwidth]{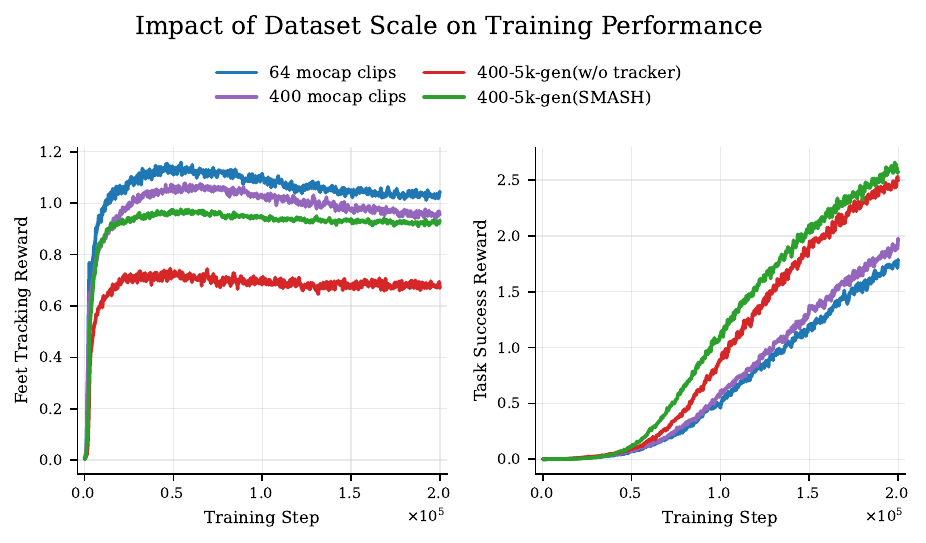} 
\caption{Impact of dataset scale on training performance.}
  \label{fig:training_curve}
\end{figure}
\begin{figure}[t!]
  \centering
  \includegraphics[width= 0.47\textwidth]{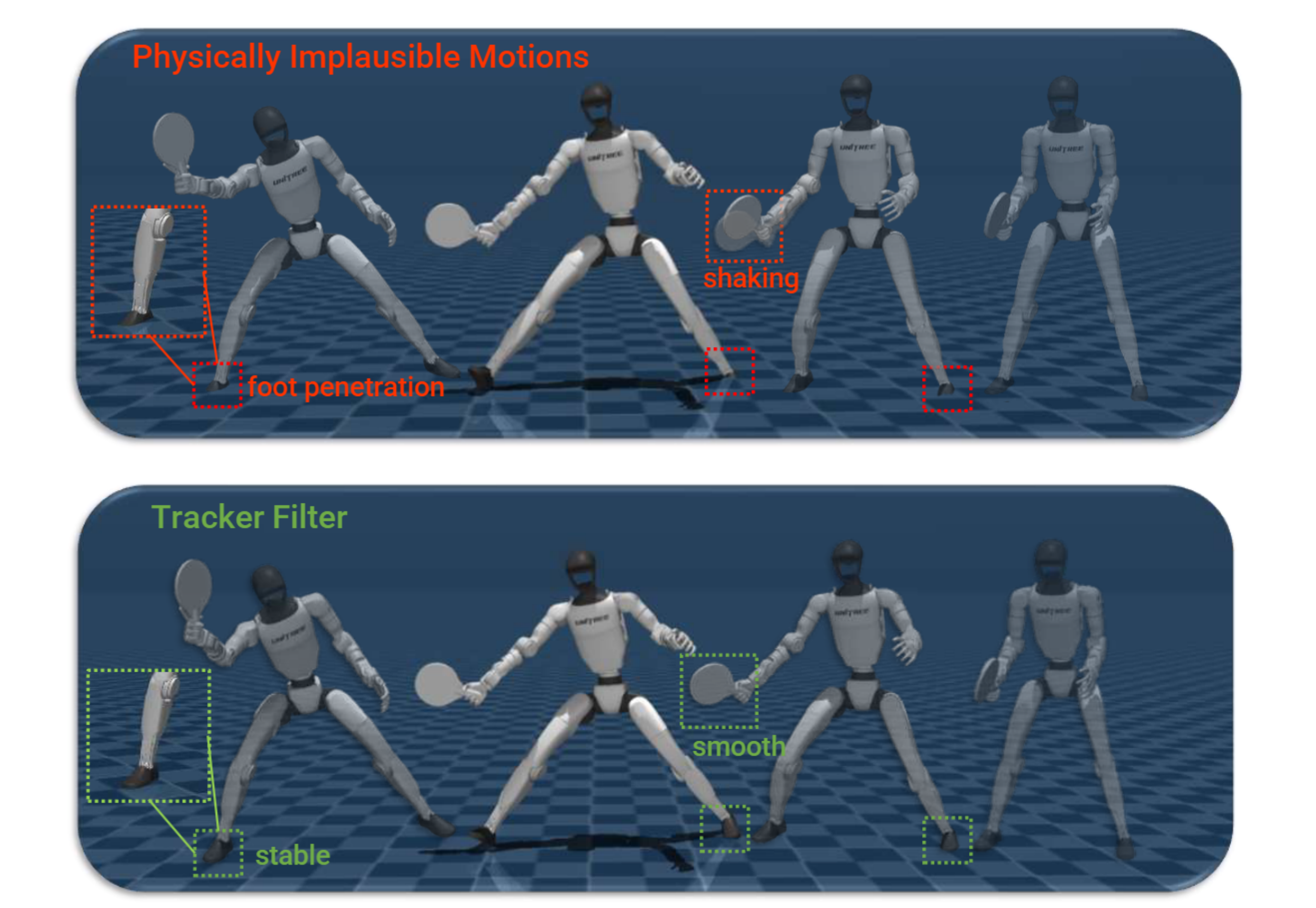} 
\caption{Comparison between physically implausible strike motions (top) and motions corrected by Tracker Filter (bottom).}
  \label{fig:tracker filter}
\end{figure}

\paragraph{Baseline Comparisons} \cref{tab:Simulation Results} provides a comprehensive evaluation of the learned policies in simulation, from three perspectives: \emph{task performance}, \emph{motion tracking accuracy}, and \emph{motion smoothness}. Overall, the results show that \textbf{ \methodname achieves the best trade-off between task execution and motion quality.} Compared with the task-only \textit{PPO} baseline, our method improves the success rate and reduces the position tracking error, indicating that motion guidance is beneficial rather than restrictive. Although PPO does not suffer from potential conflicts between motion imitation and task objectives, the lack of structured motion priors makes policy learning less effective and leads to inferior overall task performance. Compared with \textit{HITTER}, \methodname achieves comparable task success while significantly improving motion tracking accuracy and reducing torque. This is because HITTER only imitates upper-body forehand and backhand swings without lower-body tracking, so its motion space is limited and its whole-body motion consistency remains relatively poor despite the simplified motion setting. Finally, \textit{Mimic} achieves the lowest motion tracking and smoothness errors as a pure imitation baseline, verifying the accuracy of our motion tracking formulation while also showing that motion imitation alone is insufficient for solving the table-tennis task. This observation may also provide useful guidance for improving the motion-VAE module with additional dynamic constraints.

\begin{figure*}[h!]
  \centering
  \includegraphics[width= \textwidth]{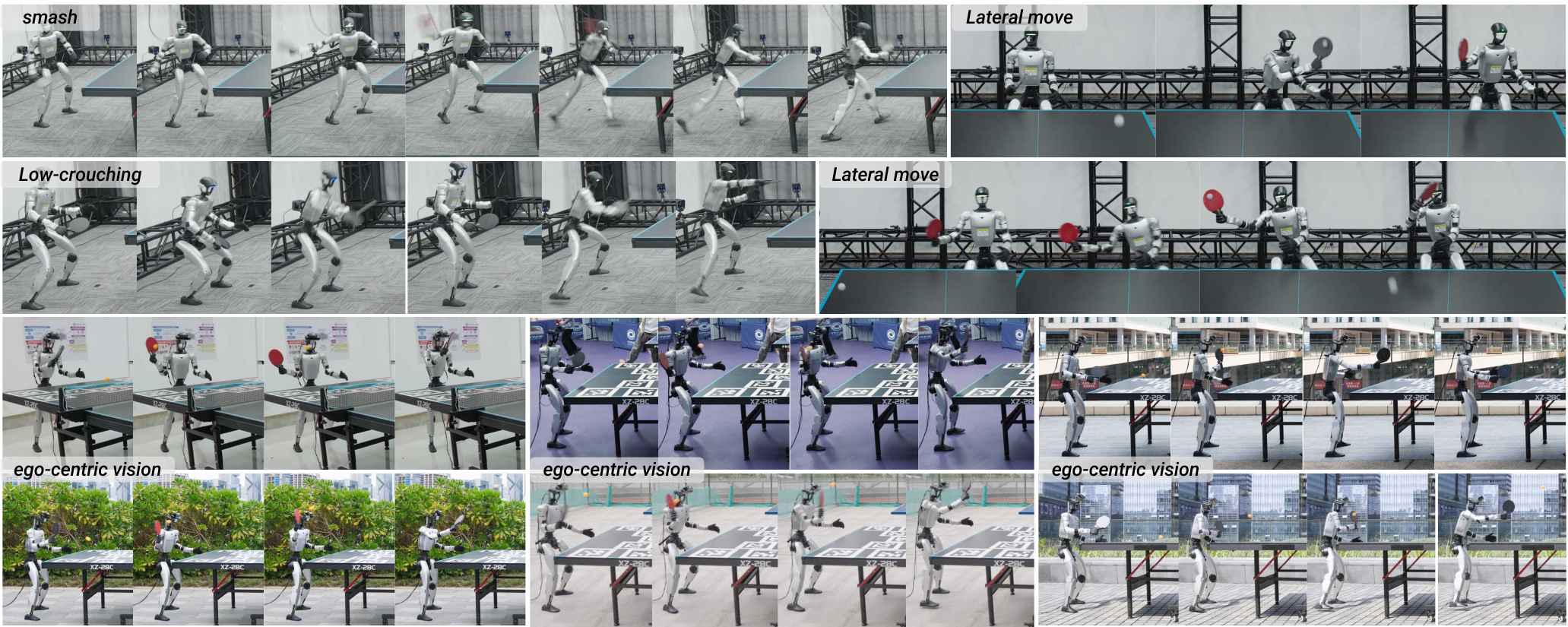} 
    \caption{\textbf{Hardware experiment snapshots.} Representative real-world snapshots of our humanoid table-tennis system, showing diverse whole-body skills such as smash, low-crouching shots, lateral movements, and ego-centric onboard strikes.}
  \label{fig:hardware}
\end{figure*}

\paragraph{Impact of MVAE Data Scaling on Policy Performance} \cref{fig:mvae_generalization} provides a direct visualization of the strike-target distribution over the table under different data scales. After motion-VAE augmentation, the strike targets become much more broadly distributed, indicating that the augmented dataset covers the hitting workspace more effectively than the original mocap set. The data-scale ablation in \cref{tab:Simulation Results} further studies how the amount of motion data affects policy learning under the same training setting. The results show that \textbf{enlarging the dataset significantly improves policy performance, confirming the importance of broad strike-motion coverage.} In particular, the policies trained with only 64, 200, or 400 mocap clips achieve substantially lower success rates than those trained with motion-VAE-generated datasets. This trend is also consistent with \cref{fig:training_curve}, where \textit{400-5k-gen} exhibits a clearly faster increase in success reward than the models trained with only 64 or 400 mocap clips. We also observe that when the initial motion set is too small, the final policy remains weaker, as shown by the lower performance of \textit{60-5k-gen} compared with \textit{200-5k-gen} and \textit{400-5k-gen}. Moreover, \textit{400-5k-gen (w/o tracker)} yields substantially worse motion tracking accuracy, especially in MPJPE, suggesting that motions generated without tracker guidance are less compatible with the robot dynamics (\cref{fig:tracker filter} ) and therefore harder for the robot to follow. This is also reflected in the lower feet-tracking reward in \cref{fig:training_curve}. On the other hand, the improvement from \textit{400-5k-gen} to \textit{400-10k-gen} is limited, indicating that once the generated motions are already broadly distributed over the reachable hitting space, further increasing the data scale brings only marginal gains.

\paragraph{Detailed Training Techniques}
We further study two training techniques: adaptive region sampling and adaptive tracking sigma. Removing adaptive region sampling causes a moderate drop in success rate, mainly due to poor performance on high strike targets above $1.2\,m$. We attribute this to the scarcity of such motions in the original dataset, since the size mismatch between human demonstrations and the humanoid robot tends to shift retargeted strike targets downward. Adaptive region sampling mitigates this issue by increasing the sampling probability of underperforming regions, thereby improving policy accuracy in these harder parts of the workspace. Adaptive tracking sigma plays a critical role in policy training. Removing this component leads to a substantial reduction in success rate and significantly degrades task-tracking performance. The reason is that an overly strict tracking sigma at the beginning makes task rewards difficult to obtain, which prevents effective task learning. 

\paragraph{Hardware Performance}
The first two rows of \cref{fig:hardware} illustrate the performance of the proposed policy on real hardware under motion-capture-based perception. Given near-ground-truth estimates of both the robot body state and the ball state from the mocap system, the policy is able to execute diverse whole-body behaviors, including smashes, crouching saves for low balls, and large-range lateral movements with coordinated whole-body motion. These results demonstrate that the proposed whole-body motion-matching mechanism can be successfully transferred to the real robot and significantly enriches the expressiveness of the robot's behaviors. We further observe that the policy can accommodate motions with substantially different styles. For example, the crouching-save behavior is learned by training jointly on a small set of crouching motions (20 clips) together with a large set of regular striking motions (5000 clips). The robot is able to perform both crouching and regular standing returns without sacrificing either behavior. We attribute this to the fact that the motion style is implicitly specified by the target ball position: when the ball arrives at significantly different locations, the policy naturally selects the corresponding motion pattern. Since smashes require a longer execution horizon, we train a separate policy with dedicated motion data for that behavior. Nevertheless, all of these results consistently show that our framework enables the humanoid robot to realize diverse whole-body motions for strikes across the full reachable workspace.

\subsection{Analysis of the Egocentric Perception System}

\begin{figure*}[t!]
  \centering
  \includegraphics[width= \textwidth]{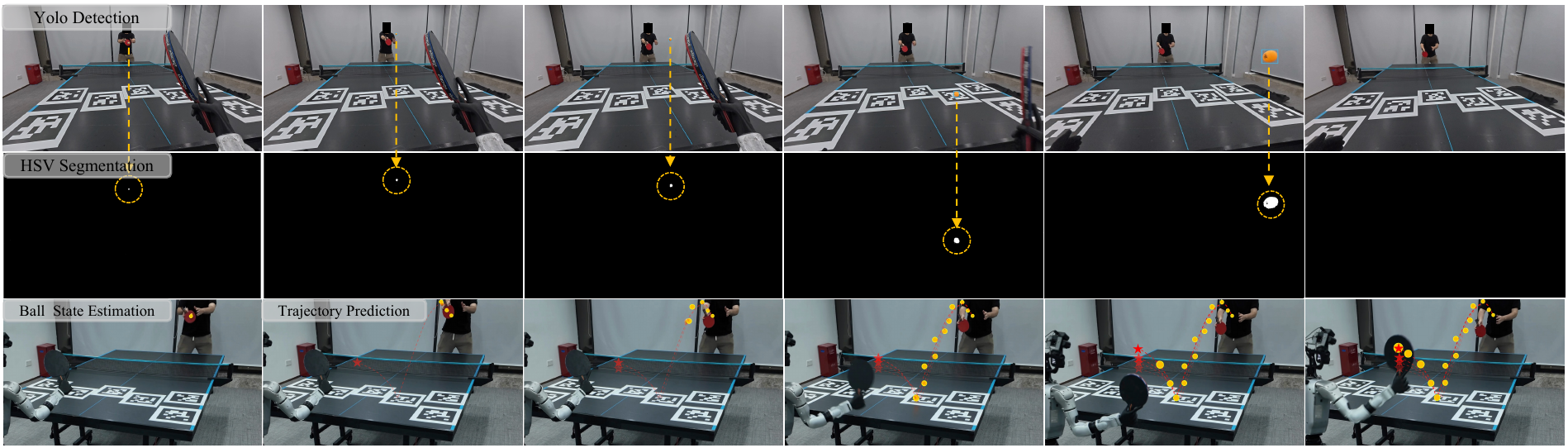} 
\caption{\textbf{Perception snapshots.} Real-time examples of onboard ball detection, HSV segmentation, ball-state estimation, and trajectory prediction during play.}
  \label{fig:perception snapshot}
\end{figure*}

\begin{figure*}[t!]
  \centering
  \includegraphics[width= \textwidth]{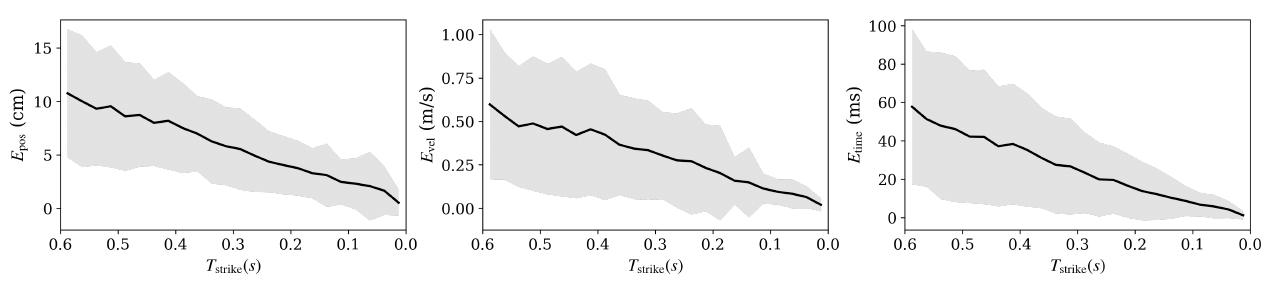} 
\caption{\textbf{Egocentric Perception Noise.} The plots show the error distribution of position (left), velocity (center), and temporal estimation (right) as a function of remaining time to strike. The horizontal axis represents the ground-truth time-to-strike, while the vertical axis denotes the magnitude of the respective estimation errors. The solid lines represent the mean error, and the shaded areas indicate the standard deviation.}
  \label{fig:perception noise}
\end{figure*}

\begin{table*}[t]
  \centering
  \caption{Evaluations against various perception settings.}
  \label{tab:perception_comparison}
  \resizebox{\textwidth}{!}{%
  \begin{tabular}{lccccccc}
    \toprule
    \textbf{Metrics}
     & $SR_{\text{hit}}\!\uparrow$   
     & $SR_{\text{return}}\!\uparrow$
     & $E_{\text{p(cm)}}\!\downarrow$ 
     & $E_{\text{o}}\!\downarrow$
     & $E_{\text{v}}\!\downarrow$
     & $E_{\text{mpjpe}}\!\downarrow$ 
     & $E_{\text{torque}}\!\downarrow$\\     
    \midrule

    Simulation (Virtual Ball)   
    & $N/A$ & $N/A$ 
    & $4.30\!\pm\! \mathsmaller{0.02}$ 
    & $4.26\!\pm\! \mathsmaller{0.07}$
    & $0.33\!\pm\! \mathsmaller{0.00}$
    & $71.40\!\pm\! \mathsmaller{0.27}$  
    & $5.51\!\pm\! \mathsmaller{0.07}$ \\

    Hardware (Virtual Ball)  
    & $N/A$ & $N/A$ 
    & $6.63\!\pm\! \mathsmaller{2.09}$ 
    & $10.91\!\pm\! \mathsmaller{3.47}$
    & $0.92\!\pm\! \mathsmaller{0.39}$
    & $95.89\!\pm\! \mathsmaller{4.41}$  
    & $6.44\!\pm\! \mathsmaller{0.32}$ \\

    Hardware (MoCap)   
    & $20/20$ & $18/20$ 
    & $6.20\!\pm\! \mathsmaller{1.91}$ 
    & $8.70\!\pm\! \mathsmaller{4.16}$
    & $1.19\!\pm\! \mathsmaller{0.31}$ 
    & $91.19\!\pm\! \mathsmaller{8.79}$ 
    & $6.29\!\pm\! \mathsmaller{0.37}$\\

    Hardware (Ego Camera) 
    & 19/20 & 12/20 
    & $6.49\!\pm\! \mathsmaller{2.26}$ 
    & $11.17\!\pm\! \mathsmaller{4.82}$
    & $1.55\!\pm\! \mathsmaller{0.54}$
    & $107.64\!\pm\! \mathsmaller{5.06}$ 
    & $6.89\!\pm\! \mathsmaller{0.36}$\\

    \bottomrule
  \end{tabular}}
\end{table*}

We first provide a qualitative overview of the egocentric perception pipeline during a representative striking sequence (\cref{fig:perception snapshot}). 
The figure visualizes the system evolution over time within a single strike, excluding the recovery phase where ball perception is not involved.

The process can be divided into two stages: perception and planning. 
The first two rows correspond to the perception stage. The first row shows the left-eye images from the head-mounted ZED\,X camera together with YOLO-based detections, which provide coarse localization of the incoming ball. 
The second row illustrates the subsequent HSV-based refinement within each bounding box, enabling more accurate pixel-level localization under varying illumination and motion blur.
The third row corresponds to the planning stage. 
Based on the triangulated 3D ball positions from stereo observations, the system performs state estimation and trajectory prediction. 
As new observations are incorporated, the AEKF progressively refines the ball state, leading to a converging prediction of the strike point as the ball approaches the robot.

\paragraph{Evaluations against Various Perception Settings}

We begin by evaluating the system under different perception settings, as summarized in Table~\ref{tab:perception_comparison}. 
Four configurations are considered. The first two, \textit{Simulation (Virtual Ball)} and \textit{Hardware (Virtual Ball)}, evaluate the striking policy using virtual ball targets, where the desired strike position and velocity are directly provided without perception noise; accordingly, $SR_{\text{hit}}$ and $SR_{\text{return}}$ are not applicable in these settings. 
The latter two, \textit{Hardware (MoCap)} and \textit{Hardware (Ego Camera)}, evaluate the full real-world striking process with physical ball interaction.

The metrics can be grouped into three categories. 
$SR_{\text{hit}}$ and $SR_{\text{return}}$ evaluate task-level striking performance. 
$E_{\text{p}}$, $E_{\text{angle}}$, and $E_{\text{v}}$ quantify racket execution accuracy at the strike instant, measuring the deviation of racket position, orientation, and velocity from the planned contact state. 
In hardware experiments, the racket pose is obtained using a rigid-body tracker mounted on the striking hand and tracked by the MoCap system. 
$E_{\text{mpjpe}}$ and $E_{\text{torque}}$ assess motion tracking quality and control effort during execution.

In simulation, the system achieves the lowest execution errors due to idealized state information and the absence of real-world sensing and hardware effects. 
When transferred to hardware with virtual balls, all execution metrics degrade noticeably, revealing a clear sim-to-real gap.
For real-ball experiments on hardware, MoCap-based perception yields higher $SR_{\text{return}}$ and lower execution errors than the ego-view setting. 
Nevertheless, the ego-view system still maintains a high $SR_{\text{hit}}$ and a nontrivial $SR_{\text{return}}$, demonstrating that the onboard perception pipeline provides sufficient accuracy for closed-loop striking in real-world conditions.
Despite the increase in both position- and velocity-related errors under ego-view perception, the policy retains reasonable task success across settings. 
This indicates that the task perturbations injection used during training is effective, enabling the policy to generalize across different perception conditions and tolerate sensing uncertainty.

Finally, the motion tracking metrics, $E_{\text{mpjpe}}$ and $E_{\text{torque}}$, remain relatively stable across perception settings. 
This suggests that perception noise primarily affects strike accuracy rather than the stability of whole-body motion execution.

Overall, these results reveal a clear sim-to-real gap in execution accuracy, while also showing that the proposed policy and perception design remain robust enough to support reliable real-world humanoid table-tennis interaction.

\begin{table*}[t]
  \centering
  \caption{Perception module ablation study.}
  \label{tab:perception_ablation}
  \resizebox{0.9\textwidth}{!}{
  \begin{tabular}{lcccccc}
    \toprule
    \textbf{Setting} 
    & $SR_{\text{det}}\uparrow$
    & $SR_{\text{hit}}\uparrow$ 
    & $SR_{\text{return}} \uparrow$ 
    & $E_{\mathrm{predpos}}\!\downarrow$
    & $E_{\mathrm{predvel}}\downarrow$
    & $SR_{\text{balance}}\uparrow$ \\
    \midrule

    w/o YOLO (no large yellow distractor) 
    & $19/20$
    & $18/20$ & $10/20$ 
    & $3.66\!\pm\!\mathsmaller{4.20}$ 
    & $0.64\!\pm\!\mathsmaller{1.14}$ 
    & $20/20$ \\

    w/o YOLO (with large yellow distractor) 
    & $0/20$
    & $N/A$ & $N/A$ & $N/A$ & $N/A$ 
    & $N/A$ \\

    w/o HSV
    & $20/20$
    & $16/20$ & $9/20$
    & $3.82\!\pm\!\mathsmaller{3.19}$
    & $0.54\!\pm\!\mathsmaller{1.03}$
    & $20/20$ \\

    w/o collision detection (AKF)
    & $19/20$
    & $17/20$ & $12/20$
    & $12.7\!\pm\!\mathsmaller{7.14}$
    & $3.37\!\pm\!\mathsmaller{1.81}$
    & $20/20$ \\

    w/o adaptive noise (AKF)
    & $18/20$
    & $16/20$ & $7/20$
    & $4.35\!\pm\!\mathsmaller{4.64}$
    & $0.61\!\pm\!\mathsmaller{1.07}$
    & $20/20$ \\

    Zero initialization (AKF)
    & $20/20$
    & $18/20$ & $12/20$
    & $6.07\!\pm\!\mathsmaller{8.61}$
    & $0.88\!\pm\!\mathsmaller{1.06}$
    & $20/20$ \\
    
    ZED X pose (instead of ZED X Mini)
    & $19/20$
    & $18/20$ & $8/20$
    & $4.70\!\pm\!\mathsmaller{5.30}$
    & $0.86\!\pm\!\mathsmaller{1.54}$
    & $19/20$ \\

    \midrule

    \textbf{Full Model} 
    & $\mathbf{20/20}$
    & $\mathbf{19/20}$ & $\mathbf{13/20}$ 
    & $\mathbf{3.49\!\pm\!\mathsmaller{2.94}}$ 
    & $\mathbf{0.53\!\pm\!\mathsmaller{0.91}}$
    & $\mathbf{20/20}$ \\

    \bottomrule
  \end{tabular}}
\end{table*}

\paragraph{Egocentric Perception Noise}

\cref{fig:perception noise} illustrates the real-world error distribution of the egocentric perception system---specifically in terms of position, velocity, and temporal estimation---as a function of the remaining time to strike. The horizontal axis represents the ground-truth time remaining until the strike event, while the vertical axis denotes the magnitude of the estimation error. Due to the limited field of view and the absence of full-trajectory ground truth, we adopt the final observed ball state prior to its exit from the camera's FOV as the reference. This serves as a reliable proxy given the significantly reduced sensing uncertainty at close range.

A consistent convergent trend is observed across all metrics: as the time to strike decreases, the estimation errors for position, velocity, and timing progressively diminish. Crucially, both the mean error and the standard deviation (indicated by the shaded areas) narrow significantly as the ball approaches the robot. This reduction in both magnitude and variance indicates that the perception system's predictions become increasingly precise and certain as the impact moment nears. This behavior aligns with the physical nature of onboard sensing, where proximity to the target enhances the signal-to-noise ratio and minimizes cumulative tracking drift, ensuring highly reliable state estimation for the final striking execution.

\paragraph{Module Ablation Study}

We conduct a module-wise ablation study to quantify the contribution of each perception component to the overall system performance. 
All experiments are performed on the ego-view pipeline with 20 trials per setting, and the results are summarized in Table~\ref{tab:perception_ablation}.

We evaluate the system using both task-level and state-estimation metrics, as summarized in Table~\ref{tab:metrics}. 
$SR_{\text{det}}$, $SR_{\text{hit}}$, and $SR_{\text{return}}$ capture end-to-end task performance, corresponding to successful detection, strike, and valid return, respectively. 
$SR_{\text{balance}}$ further evaluates post-strike stability, reflecting the sensitivity of whole-body control to perception reliability.

Estimation accuracy is quantified by $E_{\mathrm{pred_pos}}$ and $E_{\mathrm{pred_vel}}$, defined as the mean error of the predicted strike position and velocity over all prediction steps within a single strike.
Following the previous analysis, we use the last observed ball state before leaving the camera view as the reference.

The full pipeline, comprising YOLO-based coarse detection, HSV-based refinement, and the AEKF with collision handling and adaptive observation noise, achieves the best overall performance across all metrics. 
Removing the YOLO detector exposes the system to severe degradation under visual ambiguity. 
Without distractors, the system maintains a high $SR_{\text{det}}$ ($19/20$) and reasonable $SR_{\text{hit}}$, while $SR_{\text{return}}$ drops to $10/20$, indicating reduced robustness. 
Under large yellow distractors, detection fails completely ($0/20$), rendering the pipeline inoperable. 
This confirms that YOLO serves as a critical gating module for reliable candidate generation in cluttered environments.
Disabling HSV refinement preserves detection performance ($20/20$), but degrades downstream accuracy, with reduced $SR_{\text{hit}}$ and $SR_{\text{return}}$ ($16/20$ and $9/20$) and increased estimation errors. 
This indicates that coarse bounding boxes alone are insufficient, and that pixel-level refinement is necessary to suppress spatial jitter and improve measurement consistency.

Removing collision handling in the AEKF leads to the most significant degradation, with large increases in position and velocity errors ($E_{\mathrm{pred_pos}}\approx 12.7$\,cm, $E_{\mathrm{pred_vel}}\approx 3.37$\,m/s). 
This demonstrates the necessity of explicitly modeling table impacts, as bounce events introduce discontinuities that cannot be captured by smooth motion models.
Similarly, disabling adaptive observation noise reduces both estimation accuracy and task success, particularly for long-range observations. 
This validates the distance-aware uncertainty model, which mitigates overconfident updates under noisy stereo triangulation.
Using zero initialization instead of informed priors increases both temporal and state estimation errors, highlighting the importance of proper initialization for stable filtering.

Finally, replacing the ZED\,X Mini with the ZED\,X for pose estimation results in reduced $SR_{\text{return}}$ and degraded $SR_{\text{balance}}$. 
This is primarily because the head-mounted ZED\,X prioritizes long-range ball observation, leading to frequent loss of AprilTag visibility on the table and reduced pose update frequency, which can fall below the control rate. 
In contrast, the dedicated downward-facing ZED\,X Mini provides stable, high-frequency localization, thereby improving both motion stability and overall system robustness.

Overall, these results demonstrate that the perception pipeline is hierarchically structured and tightly coupled: coarse detection ensures robustness, fine localization improves measurement quality, and model-based filtering stabilizes state estimation under highly dynamic conditions. 
$SR_{\text{balance}}$ further highlights the importance of decoupled sensing, validating the effectiveness of the dual-camera design.

\begin{figure*}[t!]
  \centering
  \includegraphics[width= \textwidth]{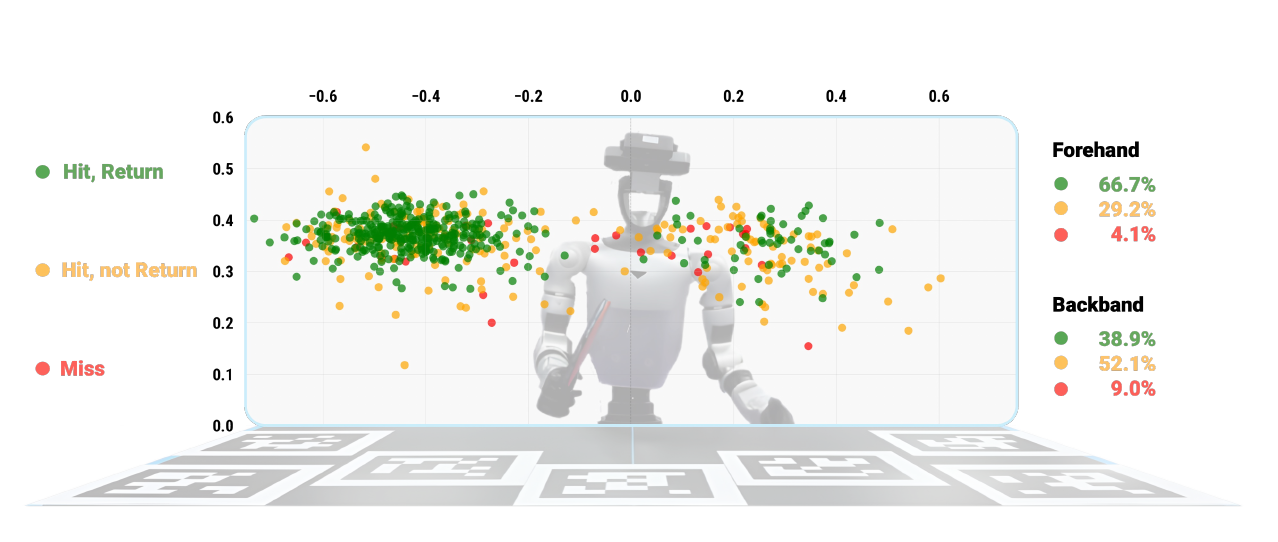} 
\caption{\textbf{Strike-point distribution and return performance.} Forehand and backhand strike points are visualized together with the corresponding hit-and-return statistics across the reachable hitting workspace.}
  \label{fig:system}
\end{figure*}

\paragraph{Robustness Analysis}

To evaluate long-term stability, we conduct a continuous 50-minute hardware experiment consisting of 642 consecutive ball launches with randomized landing locations. This setup stresses the perception system under diverse trajectories, distances, and interaction patterns. 

Across the entire dataset ($N=642$), the system achieves a total contact rate of 93.7\% (comprising a 34.0\% contact-only rate and a 59.7\% successful return rate), while only 1.2\% of trials failed due to lack of observation. Further analysis reveals a performance asymmetry between forehand and backhand strikes. The forehand achieves a return rate of 66.7\%, significantly outperforming the backhand's 38.9\%. Moreover, the backhand exhibits a notably higher miss rate (9.0\% vs. 4.1\%) and a high proportion of "contact without return" (52.1\%), indicating degraded strike quality.

This gap is primarily attributed to two factors. First, backhand motions more frequently occlude the camera field of view, leading to intermittent perception loss. Second, backhand strikes often require extended reach through aggressive posture adjustments, which introduce additional control instability and velocity tracking errors.

\section{Conclusion}

In this work, we presented \methodname, a comprehensive system for humanoid table tennis that integrates egocentric onboard perception, scalable motion generation, and task-aligned whole-body policy learning. 
Our approach enables agile and precise ball interaction with coordinated whole-body behaviors, and achieves, to the best of our knowledge, the first outdoor humanoid table-tennis demonstrations without relying on external cameras or motion-capture systems. 
These results mark a significant step toward open-world humanoid--object interaction with humanoid robots.

Despite these advances, several limitations remain. 
First, the current egocentric perception system is constrained by the limited field of view of onboard cameras. 
As a result, highly dynamic motions such as low-crouching shots and explosive smashes are difficult to deploy reliably in the egocentric setting, highlighting the need for active perception that adapt the viewpoint during interaction. 
Second, our system does not explicitly model ball spin, which is difficult to observe from visual input alone. 
Handling spin and complex serves therefore requires reactive strategies based on opponent behavior, and remains an open challenge for future work.

\clearpage
\newpage
{
\bibliographystyle{IEEEtran}
\bibliography{bibliography_short, references}
}

\end{document}